%% file: main.tex
\documentclass{article}

\usepackage[final]{corl_2022} %

\usepackage{amssymb}
\usepackage{latexsym}
\usepackage{amsmath}
\usepackage{bm}
\usepackage{amsfonts}       %
\usepackage{stmaryrd}
\usepackage[mathcal]{euscript} %

\usepackage{caption} 
\usepackage[table,dvipsnames]{xcolor} %
\usepackage{soul}
\usepackage[shortlabels]{enumitem}

\usepackage{tikz}
\usetikzlibrary{positioning}

\usepackage{pifont}
\newcommand{\cmark}{\ding{52}}%
\newcommand{\xmark}{\ding{55}}%

\usepackage{ctable}
\usepackage{booktabs,siunitx} %
\usepackage{makecell}
\usepackage{multirow}
\usepackage{subcaption} %
\usepackage{verbatim}

\usepackage{hyperref}
\hypersetup{
     colorlinks = true,
}
\usepackage{url}

\usepackage{xcolor}         %

\newcommand{\abh}[1]{\textcolor{teal}{#1}}

\newcommand{\bev}{bird's-eye-view}

\newcommand{\updated}[1]{#1}%

\makeatletter
\DeclareRobustCommand\onedot{\futurelet\@let@token\@onedot}
\def\@onedot{\ifx\@let@token.\else.\null\fi\xspace}

\def\eg{\emph{e.g}\onedot}

\makeatother

\renewcommand{\eg}{e.g., }

\makeatletter
\newcommand*{\@rowstyle}{}
\newcommand*{\rowstyle}[1]{%
 \gdef\@rowstyle{#1}%
 \@rowstyle\ignorespaces%
}
\newcolumntype{=}{%
>{\gdef\@rowstyle{}}%
}
\newcolumntype{+}{%
>{\@rowstyle}%
}

\providecommand{\paragraph}{}
\renewcommand{\paragraph}{%
  \@startsection{paragraph}{4}{\z@}%
                {1.5ex \@plus 0.5ex \@minus 0.2ex}%
                {-1em}%
                {\normalsize\bf}%
}
\providecommand{\subparagraph}{}
\renewcommand{\subparagraph}{%
  \@startsection{subparagraph}{5}{\z@}%
                {1.5ex \@plus 0.5ex \@minus 0.2ex}%
                {-1em}%
                {\normalsize\bf}%
}

\title{LaRa: Latents and Rays for Multi-Camera Bird's-Eye-View Semantic Segmentation}

\author{
  Florent Bartoccioni\\
  Valeo.ai~~~~~~~Inria\thanks{Univ.\ Grenoble Alpes, Inria, CNRS, Grenoble INP, LJK, 38000 Grenoble, France}\\
  \And
  \'Eloi Zablocki \\
  Valeo.ai
  \And
  Andrei Bursuc \\
  Valeo.ai
  \AND
  Patrick Pérez \\
  Valeo.ai
  \And
  Matthieu Cord \\
  Valeo.ai \\
  Sorbonne Université
  \And
  Karteek Alahari \\
  Inria$^*$ \\
}

\begin{document}
\maketitle

\begin{abstract}
    Recent works in autonomous driving have widely adopted the \bev\ (BEV) semantic map as an intermediate representation of the world.
    Online prediction of these BEV maps involves non-trivial operations such as multi-camera data extraction as well as fusion and projection into a common top-view grid.
    This is usually done with error-prone geometric operations (\eg homography or back-projection from monocular depth estimation) or expensive direct dense mapping between image pixels and pixels in BEV (\eg with MLP or attention).
    In this work, we present `LaRa', an efficient encoder-decoder, transformer-based model for vehicle semantic segmentation from multiple cameras.
    Our approach uses a system of cross-attention to aggregate information over multiple sensors into a compact, yet rich, collection of latent representations. %
    These latent representations, after being processed by a series of self-attention blocks, are then reprojected with a second cross-attention in the BEV space.
    We demonstrate that our model outperforms the best previous works using transformers on nuScenes. 
    The code and trained models are available at \url{https://github.com/valeoai/LaRa}.
\end{abstract}

\keywords{bird's eye view semantic segmentation; encoder-decoder transformers}

\section{Introduction}

To plan and drive safely, autonomous cars need accurate 360-degree perception and understanding of their surroundings from multiple and diverse sensors, e.g., cameras, RADARs, and LiDARs. %
Most of the established approaches tardily aggregate independent predictions from each sensor \citep{SMOKE, OFT, FCOS3D}. 
Such a late fusion strategy has limitations for reasoning globally at the scene level and does not take advantage of the available prior geometric knowledge that links sensors. %
Alternatively, the \bev's (BEV) representational space, a.k.a.\ top-view occupancy grid, recently gained considerable interest within the community.
BEV appears as a suitable and natural space to fuse multiple views~\cite{liftsplat, fiery} or sensor modalities~\cite{hendy2020fishing, bai2022transfusion} and to capture semantic, geometric, and dynamic information.
Besides, it is a widely adopted choice for downstream driving tasks including motion forecasting \citep{fiery,waymo,argoverse,nuscenes2019} and planning \citep{neuronal_planner,MP3,NEAT,nuplan}. 
In this paper, we focus on BEV perception from multiple cameras.
The online estimation of BEV representations is usually done by:
(i) imposing strong geometric priors such as a flat world~\citep{Cam2BEV} or correspondence between pixel columns and BEV rays~\citep{PON},
(ii) predicting depth probability distribution over pixels to lift from 2D to 3D and project back in BEV~\citep{liftsplat,fiery}, a system subject to compounding errors, or,
(iii) learning a costly dense mapping between multi-camera features and the BEV grid pixels \citep{CVT}.

Here, we depart from these dominant strategies and introduce `LaRa', a novel transformer-based model for vehicle segmentation from multiple cameras.
In contrast to prior works, we propose to use a latent `internal representation' instantiated as a collection of vectors.
Fusing multiple views into a compact latent space comes with several benefits.
First, it provides an explicit control on the memory and computation footprint of the model, instead of the quadratic scaling of the full mapping between multi-camera features and the BEV grid pixels \citep{CVT}. 
By design, the number of latents that we use is much smaller compared to the spatial resolution of the BEV grid, enabling a highly-efficient aggregation of information at the latent-level while exploiting spatial cues within and across camera views.
Moreover, we also hypothesize that discarding error-prone modules in the pipeline such as depth estimation \citep{liftsplat,fiery} can boost model accuracy and robustness.
Finally, we can directly predict at the full-scale BEV resolution bypassing noisy upsampling operations. This is infeasible, within a reasonable computational budget, for prior works restricted to coarser BEV grids as they map densely between all the image and BEV grid pixels~\citep{CVT}.
Besides, as an orthogonal contribution, we augment input features with ray embeddings that encode geometric relationships within and across images.
We show that such spatial embeddings, encoding prior geometric knowledge, help guide the cross-attention between input features and the latent vectors. %

Our approach is extensively validated against prior works on the nuScenes~\citep{nuscenes2019} dataset.
We significantly improve the performance on the vehicle segmentation task, outperforming recent high-performing models~\citep{liftsplat,CVT}.
Moreover, we show interesting properties of our cross-attention, which naturally stitches multiple cameras together.
We also perform several ablation and sensitivity studies of our architecture with respect to hyper-parameters changes.
Overall, LaRa is a novel model that learns the mapping from camera views to \bev\  for the task of vehicle semantic segmentation. 
In summary, our contributions are as follows:
\begin{itemize}[leftmargin=*]
\item We encode multiple views into a compact latent space that enables precise control on the model's memory and computation footprint, decoupled from the input size and output resolution.
    \item We augment semantic features with spatial embeddings derived from cameras' calibration parameters and show that it strongly helps the model learn to stitch multiple views together. %
    \item Our architectural contributions are validated on nuScenes where we reach new SOTA results.
\end{itemize}

\begin{figure}[t]
    \centering
    \includegraphics[width=\textwidth]{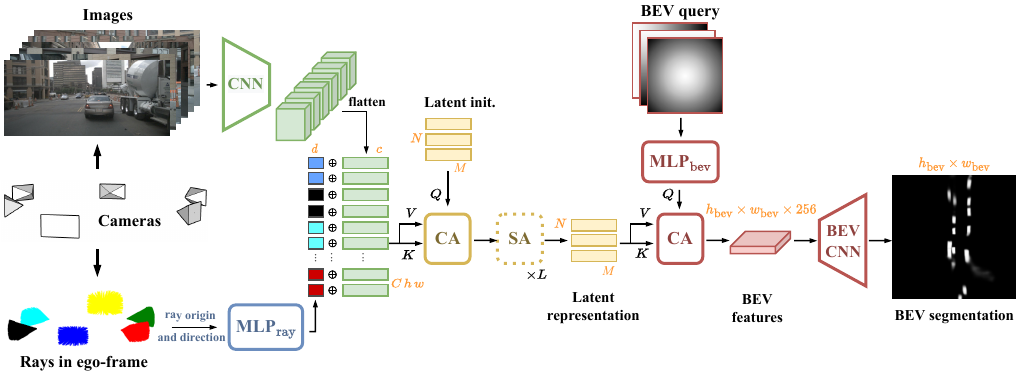}
    \caption{\textbf{LaRa overview.}
    Semantic features (green) are extracted from the images with a shared CNN and are concatenated with ray embeddings (multi-colored) that inform about geometric information to spatially relate pixels within and across cameras.
    This representation is then fused into a compact latent representation through one cross-attention (CA) and $L$ self-attention (SA) layers (yellow). The final BEV map is obtained by querying the latent representation with a cross-attention
    and then refined
    with BEV CNN (red).
    $\oplus$ denotes  concatenation. The orange letters indicate tensor dimensions. $K$, $Q$, and $V$ are the \emph{Key}, \emph{Query}, and \emph{Value} of the cross-attentions.
    }
    \label{fig:model}
\end{figure}

\section{Related work}
\label{sec:related_works}

\subsection{BEV semantic segmentation}
\label{sec:related_works:bev_semantic_segmentation}

Models for BEV segmentation are typically structured in two parts. They first extract features of each camera and then project them into a common top-view grid, called the 
\bev. There are different strategies for this projection, which can be grouped into the following categories.

\textbf{IPM-based.}
Inverse perspective mapping (IPM) defines the correspondence between the camera and the ground planes as a homography matrix. 
IPM makes strong assumptions that the world is planar and 
the cameras' horizontal axes are parallel to the ground. 
Early works 
~\citep{ARGO_IPM, IPM_BEVMapping}
apply it directly to raw camera pixels or features. 
This approach suffers from blurring and stretching artifacts for distant objects (as they have fewer pixels in the camera view) and objects with a height (as they violate the planar world assumption).
To alleviate these shortcomings, a generative adversarial network~\citep{BridgeGAN} or training a BEV decoder with synthetic ground-truth~\citep{Cam2BEV} has been used to refine the IPM projection. 

\textbf{`Lift-splat'-based: guiding with depth.}
Using depth information to lift features from 2D to 3D and then `splatting' them in BEV space recently gained popularity for its effectiveness and sound geometric definition.
Among the formulations of depth estimation for BEV projection~\citep{OFT,liftsplat,fiery,INFER,bevseg}, estimating depth probabilities along camera rays appears to perform the best~\citep{liftsplat,fiery}. 
However, such a strategy, depth being the most influential factor~\citep{MissingConfidence}, is subject to compounding errors.
Inaccuracies in depth prediction will propagate into the BEV features, which themselves can be erroneous.

\textbf{Implicitly learned with dense networks.}
An alternative to  explicit geometric projection is to learn the mapping from data.
For instance, VPN~\citep{VPN} uses an MLP to make a dense correspondence between pixels in the camera views and BEV. 
These methods rely on such expensive operations and do not use readily available spatial information given by the calibrated camera rig capturing the images.
The BEV projection 
must be entirely learned, and as it is determined by training data, it can hardly apply to new settings with slightly different camera calibrations.
Alternatively, PON~\citep{PON} builds on the observation that a column in the camera image contains all the information of the corresponding ray in BEV: it first encodes each column into a feature vector, which is then decoded into a ray along the depth dimension. 
However this relies on two implicit assumptions: (i) the camera follows a pinhole projective model, and (ii) it is horizontally aligned with the ground plane. 

\textbf{Implicitly learned with transformer architectures.}
The attention system at the core of transformer architectures~\citep{MHSA,ViT,DETR,PerceiverIO} allows learning of long-range dependencies and correspondences explicitly. %
These architectures have recently been employed for the BEV semantic segmentation task, yielding among the best-performing methods \citep{CVT,BEVFormer,GitNet}.
Nonetheless, a direct cross-attention~\citep{MHSA} between camera images and the BEV grid is computationally
expensive.
BEVFormer~\citep{BEVFormer} alleviates this issue by only cross-attending BEV pixels with cameras in which the BEV pixel is visible and by replacing the heavier multi-head attention~\citep{MHSA} with deformable attention~\citep{deformable_attention}.
CVT~\citep{CVT} keeps the vanilla multi-head cross-attention~\citep{MHSA} but applies it between low-resolution camera feature maps and a small BEV grid which is then upsampled to reach the final resolution.
GitNet~\citep{GitNet} restrains the cross-attentions to column-ray pairs making the same original implicit assumptions as PON~\citep{PON}.
Our proposed model LaRa belongs to this category as it learns the BEV representation with a transformer architecture. On the other hand, our attention scheme does not impose strong geometric assumptions while still being efficient enough to attend to a full-resolution BEV grid.

\subsection{Incorporating geometric priors in Transformers}
Since transformer architectures are permutation-invariant, spatial relationships between image regions are lost if no precautions are taken.
A standard practice to retain this spatial knowledge is to add a positional embedding to the input of attention layers~\citep{MHSA}. %
A popular approach is to encode the position of pixels with sine and cosine functions of varying frequencies~\citep{MHSA,DETR,PerceiverIO} applied over the horizontal and vertical axes.
An alternative solution to induce spatial awareness in the model is to concatenate $x,y$ positions to feature maps fed to convolutional layers~\citep{coordconv}.

Related to our ray embedding proposition, 
recent works~\citep{Inductive3DBiases, CVT}
embed the parameters of the calibrated cameras in the image features, improving training efficiency and segmentation performance. 
Similar to LaRa, IIB~\citep{Inductive3DBiases} also encodes the camera center and ray direction in the input feature sequence, but it applies it to depth estimation on image pairs in an indoor environment.
Furthermore, \citet{Inductive3DBiases} embed the origin and direction of rays into Fourier features, which can become memory intensive depending on the number of frequency bands and also introduces additional hyper-parameters to tune. 
CVT~\citep{CVT} adds up a ray direction embedding to the input feature sequence, but, differently from ours, uses the camera center embedding in the BEV query. This requires a BEV query and `cross-view attention' operation per camera, increasing the memory and computational footprint, thus limiting the maximum resolution of the BEV query.

\section{LaRa: Our Latents and Rays Model}
\label{sec:model}

Given multiple cameras observing the scene, our goal is to estimate a binary occupancy grid~\citep{Occupancygrid} ${\hat{y} \in \{0,1\}^{h_\text{bev} \times w_\text{bev}}}$ of size $h_\text{bev} \times w_\text{bev} \in \mathbb{N}^2$ for vehicles in the surroundings of the ego car. 
We propose `LaRa' a transformer-based architecture to efficiently aggregate information gathered from multiple cameras into a compact latent representation before expanding back into the BEV space.
Besides, as we believe that the geometric relationship between cameras should guide the fusion across each camera view, we propose to augment each pixel with the geometry of the ray that captured it.
The LaRa architecture is illustrated in \autoref{fig:model}.

\subsection{Input modeling with geometric priors}
\label{sec:model:input}

We consider $C$ cameras described by $(I_k, \mathcal{K}_k, \mathcal{R}_k, t_k)^C_{k{=}1}$, with $I_k \in \mathbb{R}^{H \times W \times 3}$ the image produced by camera $k$, $\mathcal{K}_k \in \mathbb{R}^{3 \times 3}$ the intrinsics, $\mathcal{R}_k \in \mathbb{R}^{3 \times 3}$ and $t_k \in \mathbb{R}^{3}$ the extrinsic rotation and translation respectively.
From these inputs, two complementary types of information are extracted: semantic information from raw images and geometric cues from the camera calibration parameters. 

\textbf{Semantic information from raw images.}
A shared image-encoder $E$ extracts feature maps for each image $F_k = E(I_k) \in \mathbb{R}^{h \times w \times c}$.
Following \citep{liftsplat, fiery},
we instantiate $E$ with a pretrained EfficientNet~\citep{EfficientNet} backbone to produce the multi-camera features.
These spatial feature maps in $\mathbb{R}^{C \times h \times w \times c}$ are then rearranged as a sequence of feature vectors, in $\mathbb{R}^{(C\,h\,w) \times c}$.

\textbf{Leveraging geometric priors.}
To enrich camera features with geometric priors, commonly used sine and cosine spatial embeddings \citep{MHSA,DETR,PerceiverIO} are ambiguous in presence of multiple cameras.
A straightforward solution would be to use camera-dependant learnable embeddings in addition to the Fourier embeddings to disambiguate between cameras. 
However, in our setting, we argue that the geometric relationship between cameras, which is defined by the structure of the camera rig, is crucial to guide the fusion of the views.
This motivates our choice to leverage the cameras' extrinsics and intrinsics to encode the position and orientation of each pixel in the vehicle ego-frame.

More precisely, we encode the camera calibration parameters by constructing the viewing ray for each pixel of the cameras.
Given a pixel coordinate $x \in \mathbb{R}^2$ within a camera image $I_k$, the direction $d_{k}(x) \in \mathbb{R}^3$ of the ray that captured $x$ is computed with:
\begin{equation}
    d_{k}(x) = \mathcal{R}^{-1}_k \mathcal{K}^{-1}_k \Tilde{x},
\end{equation}
where $\Tilde{x}$ are the homogeneous coordinates of $x$, and $d_{k}(x)$ is expressed in ego-coordinates.
The origin of the ray $d_{k}(x)$ is the camera center given by $t_k$.

Then, to fully describe the position and the orientation of the ray that captured pixel $x$, we use the embedding $\textit{ray}_k(x) \in \mathbb{R}^{d}$ computed as follows: 
\begin{equation}
\textit{ray}_k(x) = \text{MLP}_{\text{ray}}(t_k \oplus d_{k}(x)),
\end{equation}
where $\oplus$ is a concatenation operation and $\text{MLP}_{\text{ray}}$ a 2-layer MLP with GELU activations~\citep{GELU}.
The computation is consistent within and across cameras and it exhibits an interesting property: overlapping regions for two cameras with the same optical center have the same ray embedding. 
Note that the intrinsics are scaled according to the difference in resolution between $I_k$ and  $F_k$.

As shown in \autoref{fig:model}, the final input vector sequence, in $\mathbb{R}^{(C\,h\,w) \times (d + c)}$,  is produced by concatenating each of the $C\,h\,w$ feature vectors $F_k(x) \in \mathbb{R}^c$ with its geometric embedding $\textit{ray}_k(x) \in \mathbb{R}^d$. %

\subsection{Building latent representations and deep fusion}
\label{sec:model:latent}

To control the computational and memory footprint of the image-to-BEV block, we leverage findings from general-purpose architectures~\citep{PerceiverIO} and propose to use an intermediate fixed-sized latent space instead of learning the quadratic all-to-all correspondence between multi-camera features and BEV space \citep{CVT}.
Formally, the visual representations $F_k$ from all cameras, along with their corresponding geometric embeddings $\textit{ray}_k$, are compressed by cross-attention~\citep{MHSA} into a collection of $N$ learnable latent vectors of dimension $M \in \mathbb{N}$ \updated{and processed by a series of $L$ self-attention blocks~\citep{MHSA} (see yellow elements in \autoref{fig:model})}. We stress that $N \ll C\,h\,w$, which enables to fuse and process efficiently the visual information coming from all the cameras regardless of the input feature resolution or the number of cameras. 
Thanks to latent-based querying, this formulation decouples the network's deep multi-view processing from the input and output resolution. Our architecture can thus take advantage of the full resolution of the BEV grid.

\subsection{Generating BEV output from latents} %
\label{sec:model:output}

The final step is to decode the binary segmentation prediction $\hat{y} \in \{0,1\}^{h_{\text{bev}} \times w_{\text{bev}}}$ from the latent space.
In practice, the latent vectors are cross-attended~\citep{MHSA} with a BEV `query' grid $Q \in \mathbb{R}^{h_{\text{bev}} \times w_{\text{bev}} \times d_\text{bev}}$ at the final prediction resolution, with $d_\text{bev} \in \mathbb{N}$ a hyper-parameter \updated{(illustrated by the red blocks in \autoref{fig:model})}.
Each element of the query grid is a feature vector encoding the spatial position in the bird's-eye-view which specifies what information the cross-attention would extract from the latent representations.
This last cross-attention yields a feature map in BEV space, in dimension $h_\text{bev} \times w_\text{bev} \times 256$, that is further refined with a small convolutional encoder-decoder U-Net (`BEV CNN' in \autoref{fig:model}) to finally predict the binary \bev\  semantic map $\hat{y} \in \{0,1\}^{h_{\text{bev}} \times w_{\text{bev}} \times 1}$.

Specifically, we consider a combination of two types of queries: normalized coordinates in the BEV space and radial distance.  %
Normalized coordinates encode ego-centered normalized coordinates of the BEV plane.
They are obtained with: %
\begin{equation}
    Q_\text{coords}[i,j] = \left( \frac{2i}{h_{\text{bev}} - 1} - 1, \frac{2j}{w_{\text{bev}} - 1} - 1 \right),\; \forall i,j \in \{0, \dots, h_{\text{bev}} - 1\}\times\{0, \dots, w_{\text{bev}} - 1\}.
\end{equation}
Normalized radial distances are simply Euclidean distances of pixels w.r.t.\ the origin:  %
\begin{equation}
    Q_\text{radial}[i,j] = \sqrt{Q_\text{coords}[i,j]_i^2 + Q_\text{coords}[i,j]_j^2}.
\end{equation}
While the network could produce a similar embedding from $Q_\text{coords}$ using  $\text{MLP}_\text{bev}$, we find that introducing these radial embeddings along $Q_\text{coords}$ empirically improves results. Moreover, this query decoding choice compares favorably against more classical Fourier embeddings \citep{MHSA, PerceiverIO, Inductive3DBiases} and learned query embeddings \citep{MHSA,DETR}, as shown in \autoref{tab:query_ablation}.

\section{Experiments}
\label{sec:experiments}

\input{tables/prior_works}

\textbf{Dataset.}
We conduct experiments on the nuScenes dataset~\citep{nuscenes2019}, which contains 34k 
annotated sets of frames captured by $C{=}6$ synchronized cameras covering the 360° field of view around the ego vehicle.
The extrinsics and intrinsics calibration parameters are given for all cameras in every scene. 
Raw annotations come in the form of 3D bounding boxes that are simply rendered in the discretized top-down view of the scenes to form the ground-truth for our binary semantic segmentation task.

\textbf{Precise settings for training and validation.} With no established benchmarks to precisely compare model's performances, there are almost as many settings as there are previous works.
Differences are found at three different levels: The \emph{resolution} of the output grid, the \emph{level of visibility} used to select objects as part of the ground-truth, and the task considered.
In this paper, we address the task of \emph{binary semantic segmentation} of all vehicles ($\texttt{cars}$, $\texttt{bicycles}$, $\texttt{trucks}$, \textit{etc}.)~\citep{liftsplat,CVT}. This choice is made to have fair and consistent comparisons with our baselines \citep{liftsplat,CVT}, however, it should be noted that our model is not constrained to this setting.
\updated{To enable and ease future comparison, we have published our code\footnote{\url{https://github.com/valeoai/LaRa}}. We also present additional settings in the supplementary material. In all the settings we considered, models are evaluated with the IoU metric.}

\textbf{Training and implementation details.}
 We train our model by optimizing the Binary Cross Entropy
 with our predicted soft segmentation maps and the binary ground-truth.  Images are processed at resolution $224\times480$. We use the AdamW~\citep{AdamW} optimizer with a constant learning rate of $5e{-}4$ and a weight decay of $1e{-}7$. We train our model on 4 Tesla V100 16GB GPUs with a total batch size of 8 for 30 epochs. Training takes on average 11 hours. We use an EfficientNet-B4~\citep{EfficientNet} with an output stride of 8 as our CNN image encoder. For the BEV CNN we follow~\citet{liftsplat}.
 $\text{MLP}_\text{bev}$ is a 2-layer MLP producing $d_\text{bev}=128$-dimensional features.
 
\subsection{Comparison with previous works}

In \autoref{tab:sota}, we compare the IoU performances of LaRa against two baselines Lift-Splat~\citep{liftsplat} and CVT~\citep{CVT} on vehicle BEV segmentation in their respective training and evaluation setups.
In all cases, we improve results by a significant margin. More precisely, we improve by 10\% compared to Lift-Splat in their settings, by 10\% and 8\% compared to CVT respectively in Setting 1 and Setting~2. 
This suggests that our model can better extract the geometric and semantic information from all cameras with a very general architecture that does not necessitate any strong geometric assumptions.
Besides, when compared with CVT, we observe that LaRa obtains better results in the setting with finer resolution ($+10\%$ 
in Setting~1 vs.\ $+8\%$ 
in Setting 2). 

Since our attention mechanism does not rely on all-to-all attention between camera images and BEV map as CVT does,
LaRa can directly decode to the final BEV resolution which helps for fine prediction at a high resolution.

\subsection{Model ablation and sensitivity to hyper-parameters}
\label{sec:experiments:ablation}

\input{tables/query_ablation}

\textbf{Input and Output-level embeddings.}
To assess the contribution 
of the geometric embeddings that we use, we compare the different choices at both the input and output level in \autoref{tab:query_ablation}. 
As hypothesized, embedding the geometric relationship between cameras in the input is better suited for our task than the generic sine and cosine spatial embeddings. The additional camera index, while performing better than Fourier feature alone, is not enough to link pixels across cameras.
For the output query embedding, the combination of normalized coordinates and radial distance gives the best results. This simple choice outperforms both the Fourier features~\citep{MHSA, PerceiverIO} and learned embeddings~\citep{MHSA,DETR} that also have the disadvantage of increasing the number of parameters. \updated{}

\begin{figure}[t]
    \centering
    \includegraphics[width=0.9\linewidth]{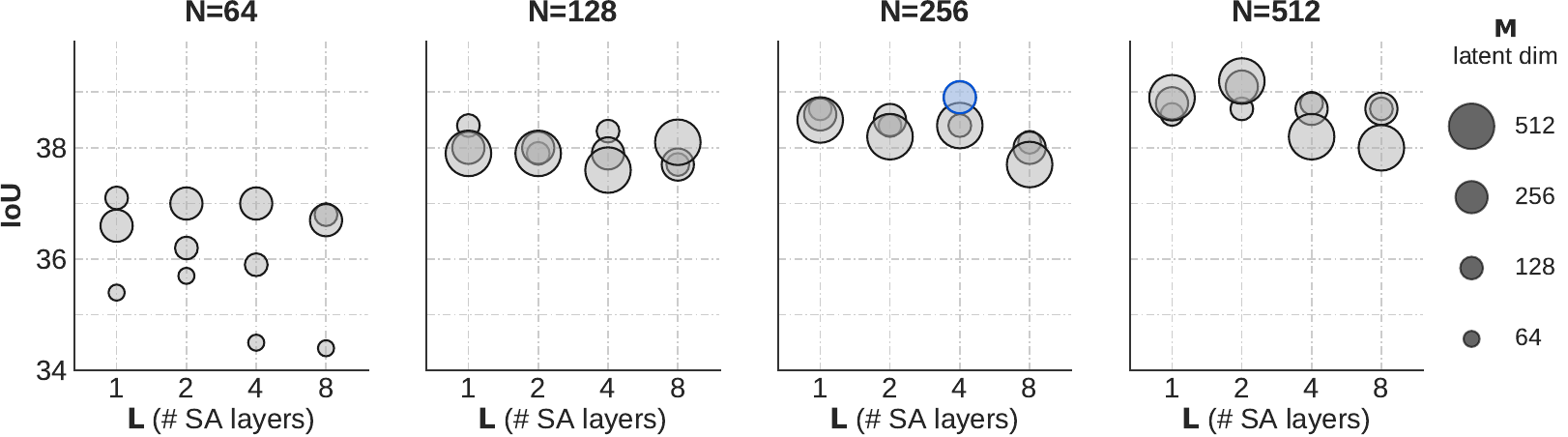}
    \caption{\textbf{Sensitivity study of LaRa to hyper-parameters.}
    We vary the number of latent vectors ($N$), their dimension ($M$), and the number of self-attention layers ($L$) and report IoU performances.
    }
    \label{fig:ablation}
\end{figure}

\textbf{Sensitivity to hyper-parameters.}
To delve into the influence of hyper-parameters, we conduct a sensitivity analysis in \autoref{fig:ablation} where we vary the number $N$ of latent vectors, their dimension $M$ and the number of self-attention blocks $L$.
We clearly observe that the performance increases with the number of latent vectors used.
This is expected as it is the main parameter controlling the attentional bottleneck between input and output.
Such a parametrization allows for an easy tuning of the performance/memory trade-off.
We observe no clear correlation between the dimension $M$ of latent vectors, the number $L$ of self-attention layers, and the obtained IoU performance. This indicates that our architecture is not too sensitive to these hyper-parameters and can work efficiently with a wide range of values for these parameters.
Although we obtain better results with 512 latent vectors, we use a maximum of 256 to stay in the same computational regime 
as
the baseline we compare against; training with 512 latent vectors requires 32GB GPUs.

\subsection{Study of attention}

As quantitatively studied in~\autoref{sec:experiments:ablation}, embedding camera rays impacts significantly the performance of LaRa. 
By analyzing the input-to-latent attention map, we further investigate the geometric reasoning of LaRa in \autoref{fig:attention_study}.
In this figure, we show two representations of the attention: a reprojection of the attention in the camera-space (left) and a top-view projection of the attention in polar coordinates by collapsing, i.e., averaging the vertical dimension (right). 
In the latter, the radial distance is proportional to the attention level and shows the directions the network attends the most.

\input{figures/Attention/attention}

The study is conducted at three different levels.
First, for a couple of one latent vector and one attention head ($n=10$, $h=5$ and $n=50$, $h=30$), among $N=256$ possible latents and $H=32$ possible attention heads.
Second, for one latent vector and the averaged attention from all attention heads ($n=10$, $h=\text{avg}$ and $n=50$, $h=\text{avg}$).
Third, for one attention head and the averaged attention over all latents ($n=\text{avg}$, $h=5$ and $n=\text{avg}$, $h=30$).
From these three settings, we 
note the followings: %
First, the attention map between one latent vector and one attention head targets a specific direction (about a 90° field of view). Additionally, it can be clearly observed that the attention is continuous across cameras, proving the network is able to retrieve the pixel relationships between views.
Second, while one attention head fires in a specific direction, the attention averaged over all the heads for one latent vector spans over half of the scene. This allows one latent vector to extract long-range context between views with the capacity to disambiguate them.
Third, the attention for one head aggregated over all the latent vectors covers all directions,  suggesting that the latent vectors contain all of the directional information and that the whole scene is attended across the latents.
To summarize, by integrating early multi-view geometric cues instantiated by camera rays embedding (\autoref{sec:model:input}), we show that LaRa learns to reason across views. \updated{We also provide quantitative evidence in the supplementary material.}

\subsection{Qualitative Results}
\label{sec:experiments:qualitative}

We show the segmentation results of two complex scenes in \autoref{fig:qualitative_study}. For a fair comparison, we use our model trained with visiblity $> 40\%$ against CVT and $> 0\%$ against Lift-Splat. Compared to LaRa, CVT missed two objects, one at a long distance and the other in the dark (red box). We also  estimate the boundaries of the vehicles better than Lift-Splat (green box). Interestingly, models trained on all vehicles (visibility $> 0\%$) tend to hallucinate cars in occluded or distant regions (highlighted with black circles in the figure).

\input{figures/qualitative/qualitative}

\section{Conclusion}
\label{sec:conclusion}

We presented LaRa, which leverages transformer-based architectures and encoder-decoder models, with respectively efficient deep cross- and self-attentions as well as an explicit control on the computation and memory footprint thanks to decoupling the bulk of the processing from the input and output resolution.
By incorporating ray embeddings into LaRa, we augment semantic features with geometric cues of the scene and show that this leads to multi-view stitching. 

\textbf{Limitations.} Our model operates on camera inputs only. Thus, in adverse conditions, e.g., with glares and darkness, its performance remains limited. To better handle these challenging situations, one avenue of improvement would be the extension of LaRa to handle complementary modalities, e.g., coming from LiDARs or radars.

\textbf{Broader impacts.} LaRa demonstrates that the geometry and semantics of a complex scene can be compacted in a small collection of latent vectors. We believe that this formulation would allow for efficient temporal reasoning. Currently, the temporal modeling is done in the BEV space, which is high resolution and mostly represents empty space~\citep{fiery,BEVFormer}.

\acknowledgments{
This work was supported in part by the ANR grants AVENUE (ANR-18-CE23-0011), VISA DEEP (ANR-20-CHIA-0022), and MultiTrans (ANR-21-CE23-0032).
It was granted access to the HPC resources of IDRIS under the allocation 2021-101766 made by GENCI.
}

\bibliography{refs} %

\input{supp}

\end{document}

%% file: tables/prior_works.tex
\begin{table}[]

\caption{\textbf{Intersection-over-Union (IoU) for vehicle segmentation on nuScenes.}
`Setting 1' refers to a 100m$\times$50m grid with a 25cm resolution and `Setting 2' to a 100m$\times$100m grid with a 50cm resolution. For training and validation, vehicles are considered only if their visibility level is above a predefined threshold (either $0\%$ or $40\%$).
To compare against other works, we refer the reader to Lift-splat \citep{liftsplat} and CVT \citep{CVT}.}
\smallskip
\label{tab:sota}
\centering
\begin{tabular}{l c c c c}
\toprule

& & \multicolumn{1}{c}{visibility $> 0\%$} & \multicolumn{2}{c}{visibility $> 40\%$} \\ 
\textbf{Method} & Conference & Setting 2 & Setting 1 & Setting 2  \\
\midrule

Lift-splat \citep{liftsplat} & ECCV'20 & 32.1 & --- & ---\\
CVT \citep{CVT} & CVPR'22 & --- & 37.5 & 36.0 \\
LaRa (ours) & --- & \textbf{35.4} & \textbf{41.4} & \textbf{38.9}\\ %

\bottomrule

\end{tabular}
\label{tab:SettingOne}
\vspace{-3pt}
\end{table}

%% file: tables/query_ablation.tex
\begin{table}[]
\addtolength{\tabcolsep}{-2.5pt}

\caption{\textbf{Ablation study for the input and output query embedding}. Training and evaluation are done in Setting 2 (100m$\times$100m at 50cm resolution), with a visibility $>0\%$.
}

\smallskip

\resizebox{\textwidth}{!}{
\begin{subtable}[t]{.4\textwidth}
 \raggedleft
 \scalebox{0.99}{
\begin{tabular}{ c c c c}
\toprule

\multicolumn{3}{c}{Input geometry embedding} &   \\ 
\cmidrule(l){1-3}
 Cam. rays & Cam. idx & Fourier & \textbf{IoU}\\
\midrule

\cmark & \xmark & \xmark & \textbf{35.4} \\
\cmark & \cmark & \cmark & 34.4 \\ %

\xmark & \cmark & \cmark & 32.3 \\
\xmark & \xmark & \cmark & 30.5 \\

\bottomrule
\end{tabular}
}
\end{subtable}%

\hspace{40pt}

\begin{subtable}[t]{.6\textwidth}
        
\scalebox{0.99}{
\begin{tabular}{c c c c c}
\toprule

\multicolumn{4}{c}{Output query embedding} &   \\ 
\cmidrule(l){1-4}
Radial dist.\ & Norm.\ coords & Fourier & Learned & \textbf{IoU}\\
\midrule

\cmark & \cmark & \xmark & \xmark & \textbf{35.4} \\ 
\xmark & \cmark & \xmark & \xmark & 35.1 \\ 
\xmark & \xmark & \cmark & \xmark &  30.6\\ 
\xmark & \xmark & \xmark & \cmark & 21.8 \\

\bottomrule
\end{tabular}
}
\end{subtable}%
}

\label{tab:query_ablation}
\vspace{-10pt}
\end{table}

%% file: figures/Attention/attention.tex
\newcommand{\camsimage}[1]{\includegraphics[width=0.55\linewidth]{#1}}
\newcommand{\polarimage}[1]{\includegraphics[width=0.15\linewidth]{#1}}

\definecolor{colorF}{HTML}{0053D6}
\definecolor{colorFL}{HTML}{D60000}
\definecolor{colorFR}{HTML}{666666}
\definecolor{colorB}{HTML}{D6D200}
\definecolor{colorBL}{HTML}{04D600}
\definecolor{colorBR}{HTML}{00D6CF}

\begin{figure*}[b]
\vspace{-5pt}
\centering
\resizebox{\textwidth}{!}{
\begin{tikzpicture}[
    every node/.style={inner sep=0,outer sep=2},
    label/.style = {
        inner sep=2pt,
        font=\scriptsize,
        align=center,
    },
    scores/.style = {
        inner sep=2pt, 
        outer sep=0pt,
        font=\footnotesize,
        text=white,
        align=center,
        anchor=north,
    },
    tag/.style = {
        fill=black,
        inner sep=1pt, 
        outer sep=0pt,
        yshift=2pt,
        xshift=2pt,
        font=\scriptsize,
        text=white,
        align=center,
        anchor=south west,
    },
    camtag/.style = {
        inner sep=1pt, 
        outer sep=0pt,
        font=\tiny,
        text=white,
        align=center,
        fill opacity=0.5,
        text opacity=1
    },
]

    \node (Cam10) {\camsimage{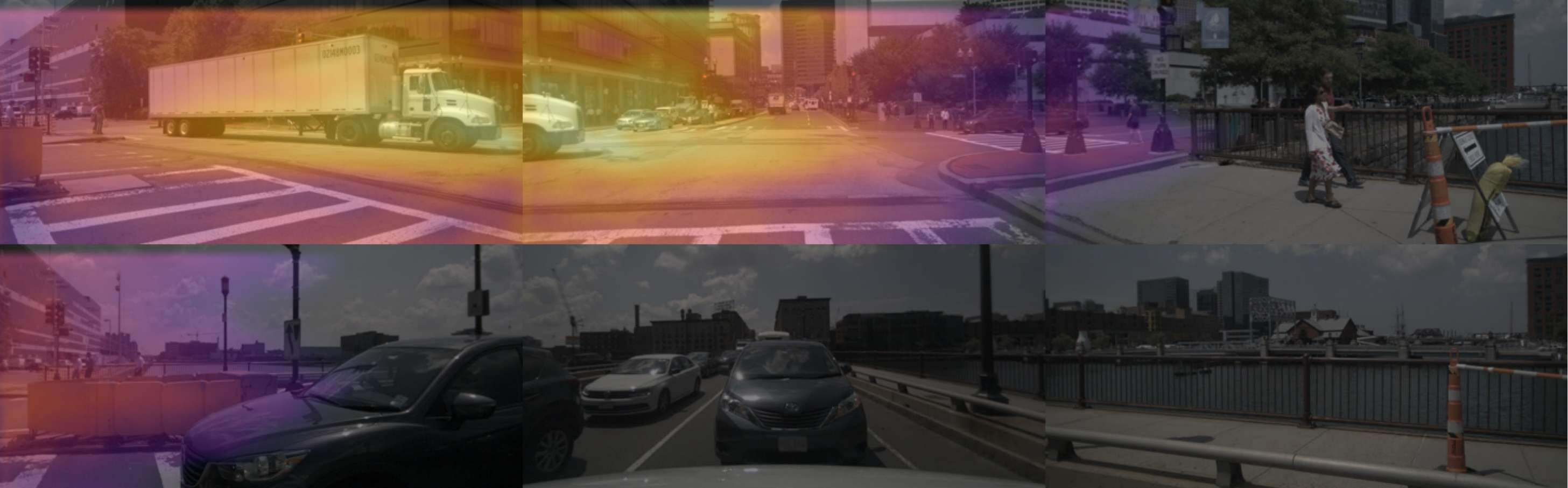}};
    \node[rectangle, tag] (Cam10_tag) at (Cam10.south west) {n:10 h:5};
    \node[rectangle, camtag, fill=colorFL,  anchor=north west, yshift=-2pt, xshift=20pt] (Cam10_FL) at (Cam10.north west) {front left};
    \node[rectangle, camtag, fill=colorF, anchor=north, yshift=-2pt] (Cam10_F) at (Cam10.north) {front};
    \node[rectangle, camtag, fill=colorFR, anchor=north east, yshift=-2pt, xshift=-20pt] (Cam10_FR) at (Cam10.north east) {front right};
    \node[rectangle, camtag, fill=colorBL, anchor=west, yshift=-3pt, xshift=20pt] (Cam10_BL) at (Cam10.west) {back left};
    \node[rectangle, camtag, fill=colorB, anchor=center, yshift=-3pt] (Cam10_B) at (Cam10.center) {back};
    \node[rectangle, camtag, fill=colorBR, anchor=east, yshift=-3pt, xshift=-20pt] (Cam10_BR) at (Cam10.east) {back right};
    
    \node [anchor=west] (Polarl10h5) at (Cam10.east) {\polarimage{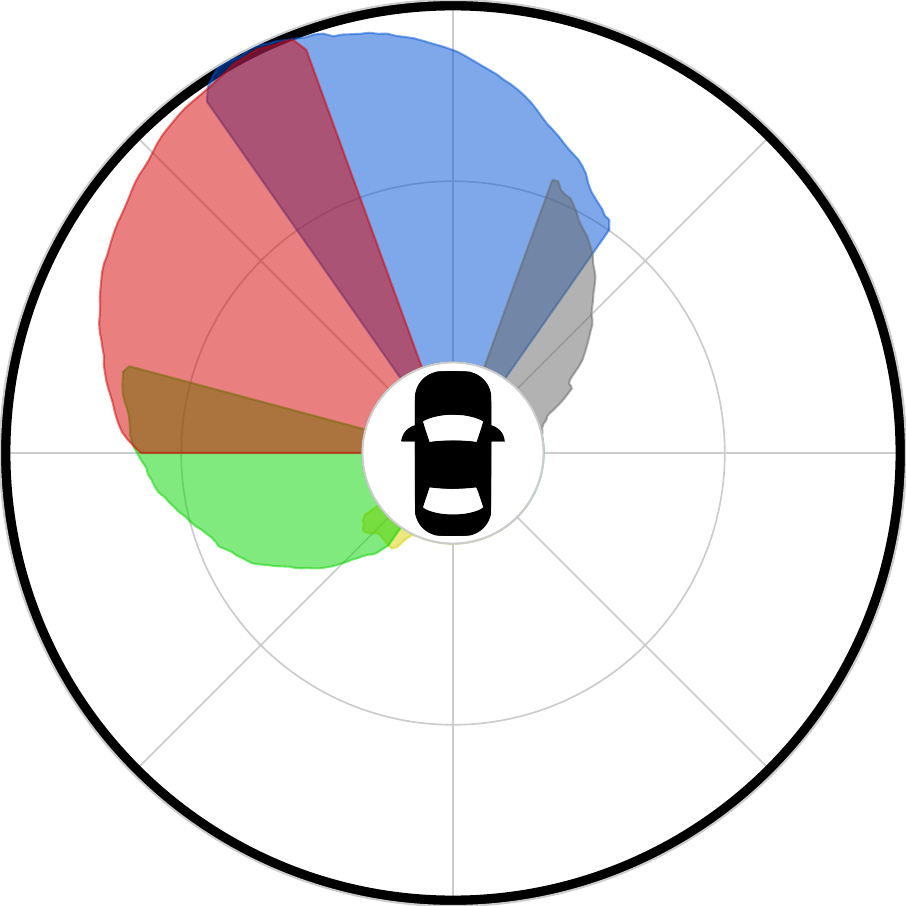}};
    \node[rectangle, tag] (Polarl10h5_tag) at (Polarl10h5.south west) {n:10 h:5};
    
    \node [anchor=west] (Polarl10havg) at (Polarl10h5.east) {\polarimage{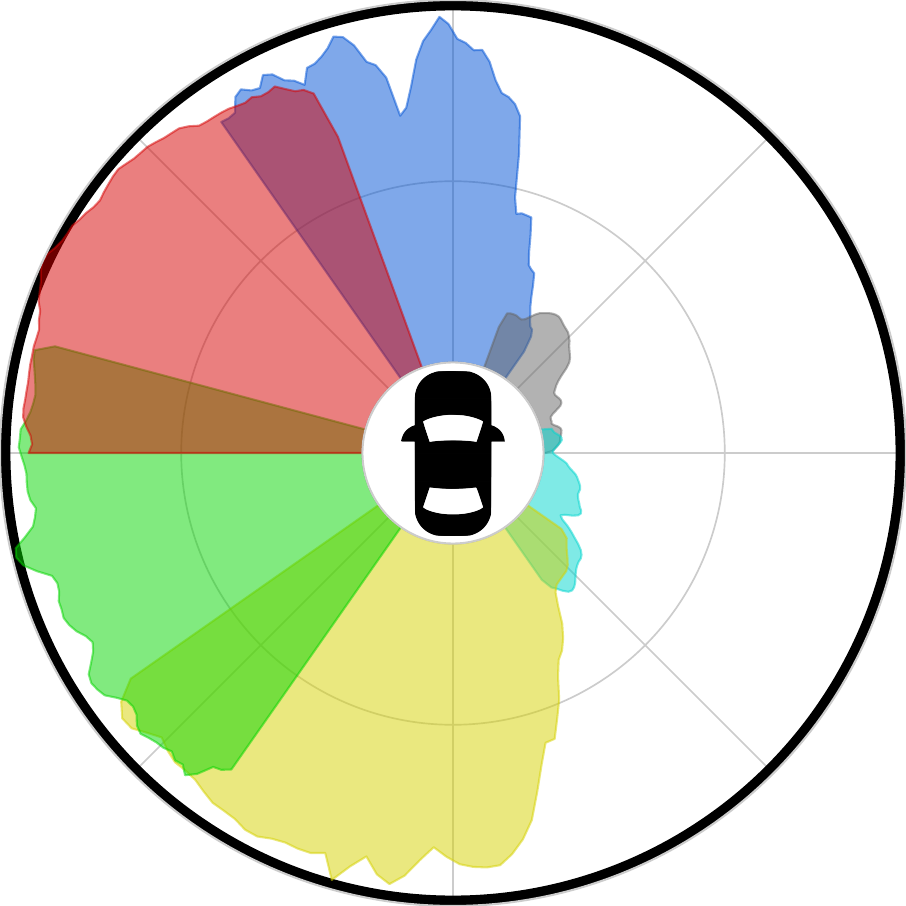}};
    \node[rectangle, tag] (Polarl10havg_tag) at (Polarl10havg.south west) {n:10 h:avg};
    
    \node [anchor=west] (Polarlavgh5) at (Polarl10havg.east) {\polarimage{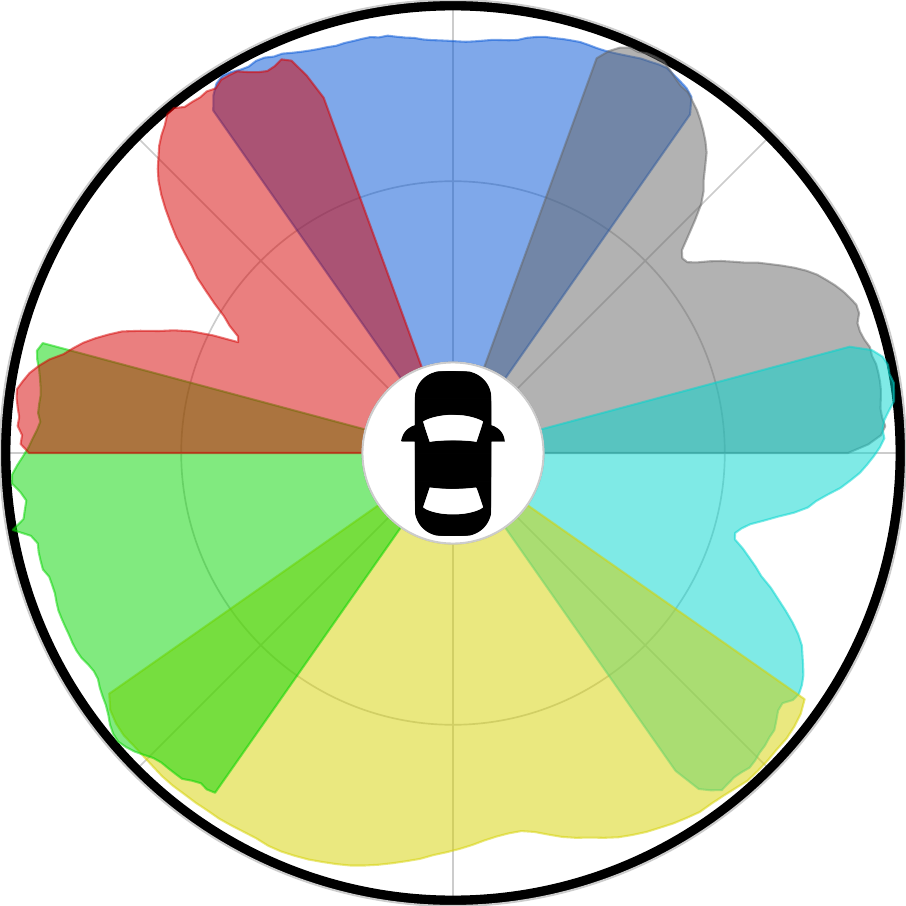}};
    \node[rectangle, tag] (Polarlavgh5_tag) at (Polarlavgh5.south west) {n:avg h:5};

    \node [anchor=north] (Cam50) at (Cam10.south) {\camsimage{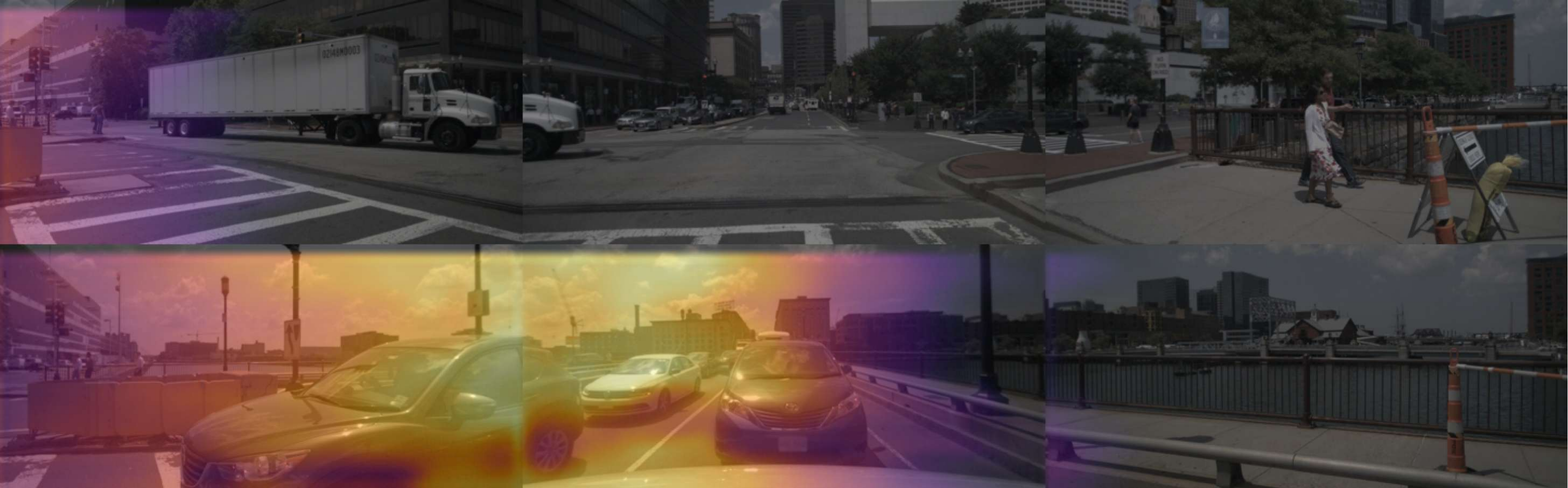}};
    \node[rectangle, tag] (Cam50_tag) at (Cam50.south west) {n:50 h:30};
    \node[rectangle, camtag, fill=colorFL,  anchor=north west, yshift=-2pt, xshift=20pt] (Cam50_FL) at (Cam50.north west) {front left};
    \node[rectangle, camtag, fill=colorF, anchor=north, yshift=-2pt] (Cam50_F) at (Cam50.north) {front};
    \node[rectangle, camtag, fill=colorFR, anchor=north east, yshift=-2pt, xshift=-20pt] (Cam50_FR) at (Cam50.north east) {front right};
    \node[rectangle, camtag, fill=colorBL, anchor=west, yshift=-3pt, xshift=20pt] (Cam50_BL) at (Cam50.west) {back left};
    \node[rectangle, camtag, fill=colorB, anchor=center, yshift=-3pt] (Cam50_B) at (Cam50.center) {back};
    \node[rectangle, camtag, fill=colorBR, anchor=east, yshift=-3pt, xshift=-20pt] (Cam50_BR) at (Cam50.east) {back right};
    
    \node [anchor=west] (Polarl50h30) at (Cam50.east) {\polarimage{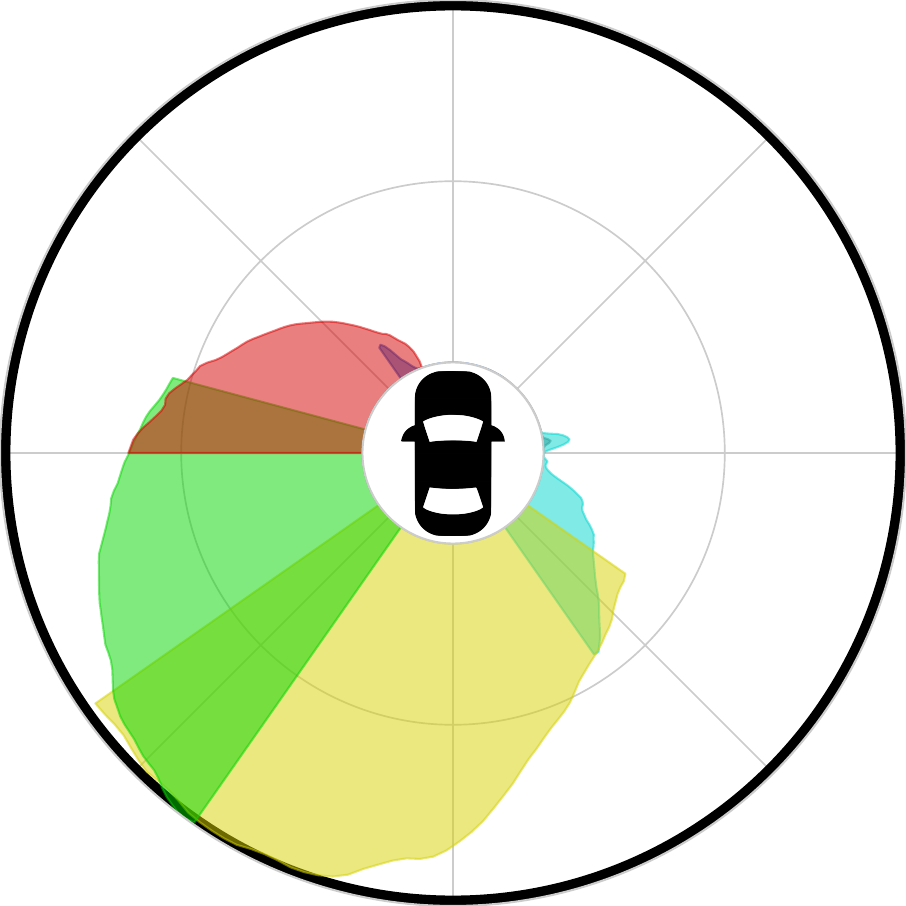}};
    \node[rectangle, tag] (Polarl50h30_tag) at (Polarl50h30.south west) {n:50 h:30};
    
    \node [anchor=west] (Polarl50havg) at (Polarl50h30.east) {\polarimage{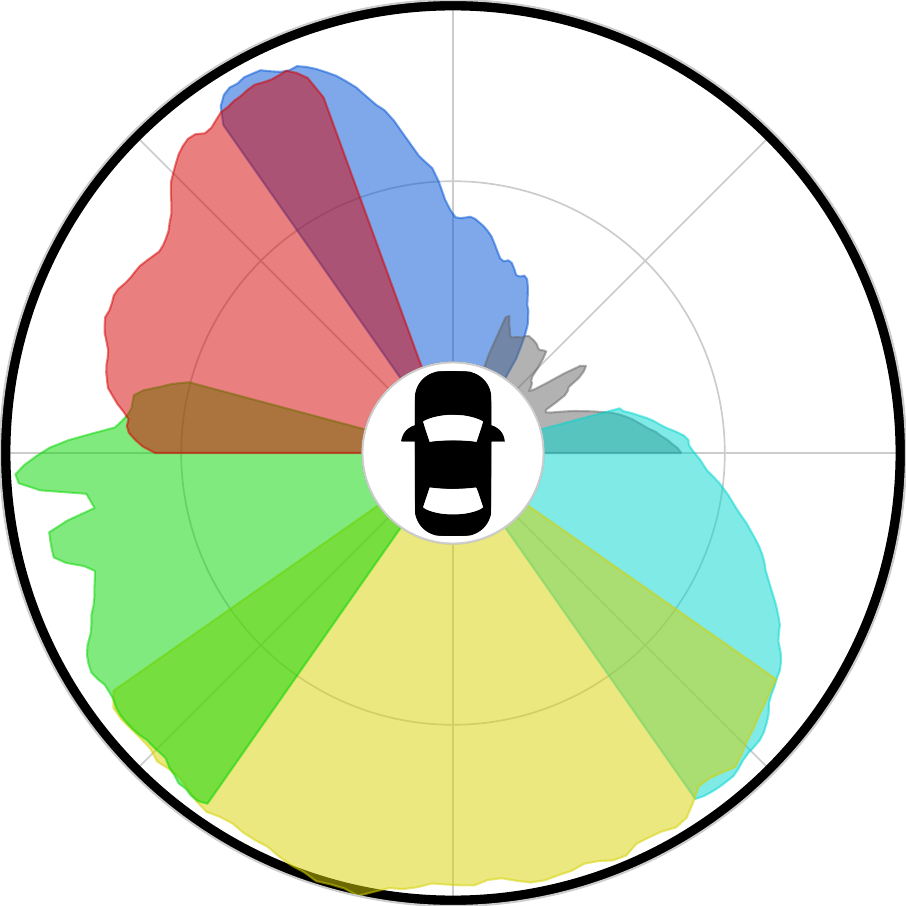}};
    \node[rectangle, tag] (Polarl50havg_tag) at (Polarl50havg.south west) {n:50 h:avg};
    
    \node [anchor=west] (Polarlavgh30) at (Polarl50havg.east) {\polarimage{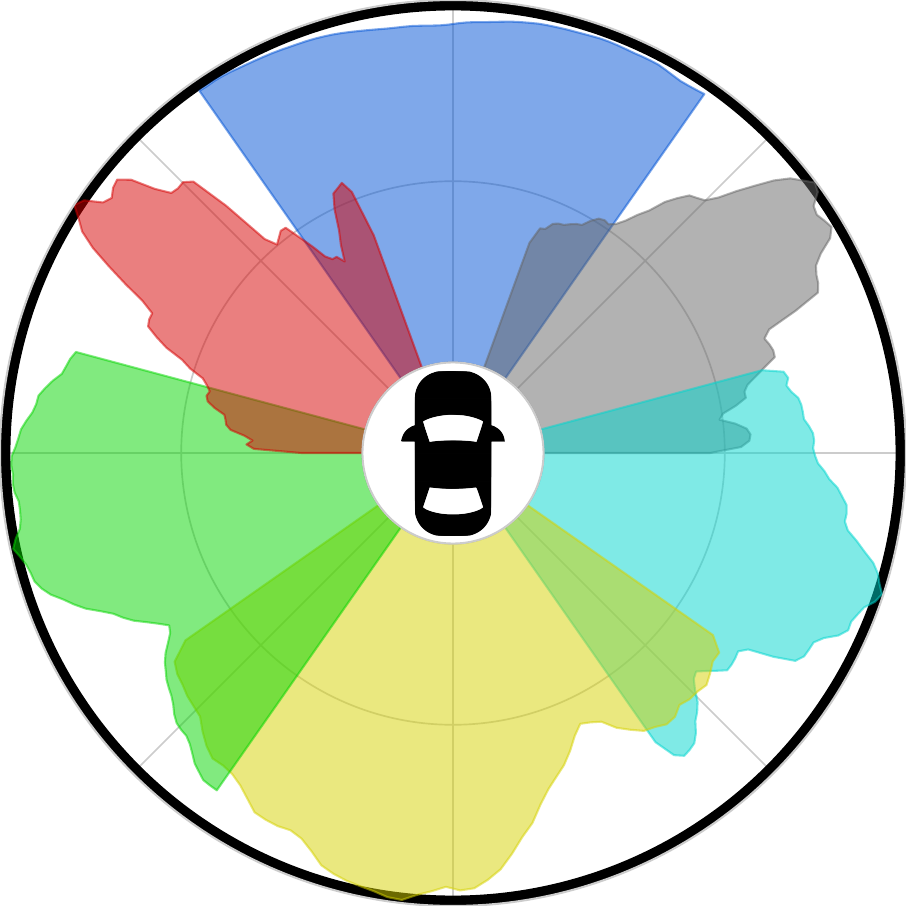}};
    \node[rectangle, tag] (Polarlavgh30_tag) at (Polarlavgh30.south west) {n:avg h:30};

\end{tikzpicture}
}

\caption{\textbf{Input-to-latent attention study.} Attention maps analysis for a network using 256 latents and 32 attention heads. The attention for one attention head and one latent is shown on the left superimposed with RGB images. The polar plots represent the directional attention intensity for one (or the average) attention head with one (or the average) latent vector. %
The radial distance is proportional to the attention level and shows the directions the network attends the most.
}
\label{fig:attention_study}
\end{figure*}
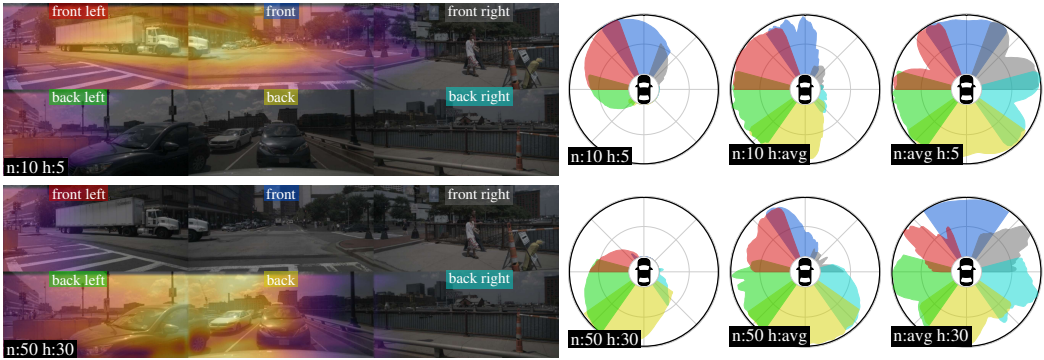

%% file: figures/qualitative/qualitative.tex
\newcommand{\camsband}[1]{\includegraphics[width=1.\linewidth]{#1}}

\newcommand{\bevimage}[1]{\includegraphics[width=0.25\linewidth]{#1}}

\begin{figure*}[t]
\centering
\resizebox{1.\linewidth}{!}{
\begin{tikzpicture}[
    tag/.style = {
        fill=black,
        inner sep=1pt, 
        outer sep=0pt,
        yshift=2pt,
        xshift=2pt,
        font=\scriptsize,
        text=white,
        align=center,
        anchor=south west,
    },
    camtag/.style = {
        inner sep=1pt, 
        outer sep=0pt,
        font=\tiny,
        text=white,
        align=center,
        fill opacity=0.5,
        text opacity=1
    },
]

    \node [inner sep=0pt, outer sep=0pt] (Cam0) at (0,0) {\camsband{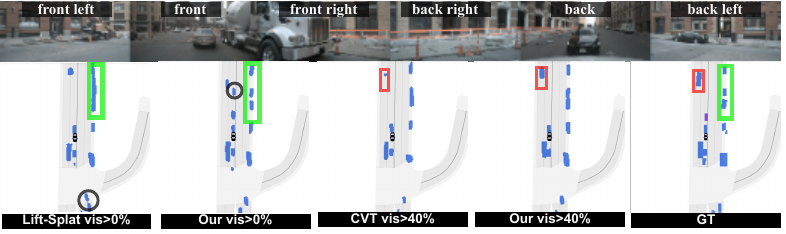}};

     \node [anchor=north] (Cam5055) at (Cam0.south) {\camsband{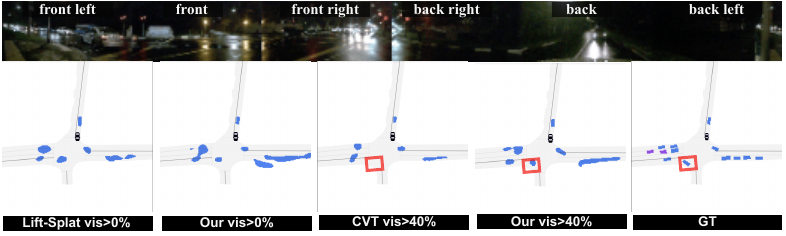}};

\end{tikzpicture}
}

\caption{\textbf{Qualitative results on complex scenes.} We show the six camera views surrounding the vehicle along with segmentation map ground-truth for reference.
In the ground-truth (GT) map, vehicles are shown in blue (visibility $>40\%$) or purple (visisibility $<40\%$). %
The ego vehicle is located in the center and facing downwards.
We show results for our two baselines~\citep{liftsplat, CVT}. For a fair comparison, we always compare using their respective level of visibility. Setting 2 is used.
}
\label{fig:qualitative_study}
\vspace{-7pt}
\end{figure*}

%% file: supp.tex
\newpage
\appendix

\part*{LaRa: Latents and Rays for Multi-Camera Bird's-Eye-View Semantic Segmentation\\ \centering --- Supplementary Material ---}

\section{Implementation details}

Following common practice~\citep{liftsplat,fiery,CVT} we employ an EfficientNet~\citep{EfficientNet} as our CNN image encoder $E$. In particular, we use an EfficientNet-B4~\citep{EfficientNet} with an output stride of 8. It extracts feature maps for each image $F_k = E(I_k) \in \mathbb{R}^{h \times w \times c}$. In practice, $h = 224 / 8 = 28$, $w = 480 / 8 = 60$ and we define $c = 128$.

For the BEV CNN, we follow~\citet{liftsplat} and use an encoder-decoder architecture with a ResNet-18~\citep{resnet} as backbone. It produces features at three levels of resolutions (1:1, 1:2 and 1:8), which are progressively upsampled back to the input resolution with bilinear interpolation (first $\times$4 for the 1:8\textit{th} scale then $\times$2 for the 1:2\textit{th}). Skip connections are used between encoder and decoder stages of the same resolution.

Both $\text{MLP}_\text{ray}$ and $\text{MLP}_\text{bev}$ are 2-layer MLPs producing 128-dimensional features. Each consists of two linear transformations with a GELU~\citep{GELU} activation function:
\begin{equation}
    \text{MLP}(x) = W_2 \text{GELU}(W_1 x + b_1) + b_2.
\end{equation}

The exact specification of other modules will be available in our code upon publication.

\subsection{Attention modules}

Following the original formulation and notations~\citep{MHSA}, the attention operation is defined as:
\begin{equation}
    \label{eq:attn}
    \text{Attn}(Q, K, V) = \text{softmax}(\frac{QK^{\top}}{\sqrt{d_K}})V
\end{equation}

 with its multi-headed extension:
\begin{equation}
    \begin{split}
    \text{MultiheadAttn}(Q, K, V) &= \text{Concat}(\text{head}_1, \dots, \text{head}_h) W^O \\
    \text{where } \text{head}_i &= \text{Attn}(QW_i^Q, KW_i^K, VW_i^V).
\end{split}
\end{equation}

with $d_q$, $d_v$, $d_k$ the dimensions of $Q$, $K$ and $V$. In practice, we use $d_\text{model}$, a hyperparameter, to define the dimension of the queries, keys and values for the inner attention~(\autoref{eq:attn}) as well as $h$ the number of attention heads. More precisely, we linearly project queries, keys and values $h$ times with different projections, each with dimension $d_\text{emb} = d_\text{model}/h$. The learnable projection matrices of each head are defined as $W_i^Q \in \mathbb{R}^{d_q\times d_\text{emb}}$, $W_i^K \in \mathbb{R}^{d_k\times d_\text{emb}}$, $W_i^V \in \mathbb{R}^{d_v\times d_\text{emb}}$ and $W_i^O \in \mathbb{R}^{h \cdot d_\text{emb} \times d_v}$.

Our architecture integrates three attention modules~\citep{MHSA}: (i) a cross-attention between latent vectors and input features; (ii) a sequence of self-attention on the latent vectors; (iii) a cross-attention between BEV query and latent vectors. More precisely, and with a slight abuse of notation:

\textbf{Latent-Input cross-attention}
\begin{equation}
\begin{split}
    \text{latents} &:= \text{MultiheadAttn}(\text{LayerNorm}(\text{latents}), \text{LayerNorm}(\text{input}), \text{LayerNorm}(\text{input})) + \text{latents} \\
    \text{latents} &:= \text{MLP}(\text{LayerNorm}(\text{latents})) + \text{latents}
\end{split}
\end{equation}
\textbf{Latent self-attention}
\begin{equation}
\begin{split}
    \text{latents} &:= \text{MultiheadAttn}(\text{LayerNorm}(\text{latents}), \text{LayerNorm}(\text{latents}), \text{LayerNorm}(\text{latents})) + \text{latents} \\
    \text{latents} &:= \text{MLP}(\text{LayerNorm}(\text{latents})) + \text{latents}
\end{split}
\end{equation}
\textbf{BEVquery-Latent cross-attention}
\begin{equation}
\begin{split}
    \text{output} &:= \text{MultiheadAttn}(\text{LayerNorm}(\text{BEVquery}), \text{LayerNorm}(\text{latents}), \text{LayerNorm}(\text{latents})) \\
    \text{output} &:= \text{MLP}(\text{LayerNorm}(\text{output})) + \text{output}
\end{split}
\end{equation}

In particular, the cross-attention between BEV query and latent vectors is not residual. 
Since the query is made of coordinates, imposing the network to predict segmentation as residual of coordinates does not make sense.

\subsection{Output embedding}

In the main paper, we considered Fourier features and learned query as alternative BEV query embeddings. Here we detail both of them.

\paragraph{Fourier features.} The Fourier encoding has been proven 
to be well suited for encoding fine positional features~\citep{MHSA, PerceiverIO, Inductive3DBiases}. This is done by applying the following on an arbitrary input $z \in \mathbb{R}$:
\begin{equation}
    \text{fourier}(z) = \left(z, \sin(f_1 \pi z), \cos(f_1 \pi z), \dots, \sin(f_B \pi z), \cos(f_B \pi z)  \right),
\end{equation}
where $B$ is the number of Fourier bands, and $f_b$ is spaced linearly from 1 to a maximum frequency $f_B$ and typically set to the input’s Nyquist frequency~\citep{PerceiverIO}.
The maximum frequency $f_B$ and number of bands $B$ are hyper-parameters.
This Fourier embedding is applied on the normalized coordinate grid such that:
\begin{equation}
    Q_\text{fourier}[i,j] = \text{fourier}(Q_\text{coords}[i,j]_i) \oplus \text{fourier}(Q_\text{coords}[i,j]_j).
\end{equation}

\paragraph{Learned.} Another alternative, following common transformer practice~\citep{MHSA,DETR} and most notably proposed by CVT~\citep{CVT}, is to let the network learn its query of dimension $d_\text{bev-query}$ from data. However, this is memory intensive as it introduces $h_{\text{bev}} \times w_{\text{bev}} \times d_\text{bev-query}$ additional parameters to be optimized.
In other words, the number of parameters grows quadratically to the resolution of the BEV map. For experiments using learned output query embedding, we use $d_\text{bev-query} = 32$.

\section{Evaluation details}

With no established benchmarks to precisely compare model's performances, there are almost as many settings as there are previous works.
Differences are found at three different levels: 

\begin{itemize}
    \item The \textbf{resolution} of the output grid where two main settings have been used: a grid of 100m$\times$50m at a 25cm resolution \citep{CVT,PON,VPN,STA} and a grid of $100$m$\times100$m at a $50$cm resolution \citep{liftsplat,CVT}. These settings are respectively referred as `Setting 1' ($h_\text{bev} \times w_\text{bev} = 400 \times 200$)  and `Setting 2' ($h_\text{bev} \times w_\text{bev} = 200 \times 200$). %
    
    \item The considered \textbf{classes}. There are slight differences in the classes used to train and evaluate the model. For instance, some models are trained with a multi-class objective to simultaneously segment objects such as \abh{$\texttt{cars}$, $\texttt{pedestrian}$ or $\texttt{cones}$}~\citep{PON,VPN,STA}.
    Some others only train and evaluate in a binary semantic segmentation setting on a meta-class \abh{$\texttt{vehicles}$} which includes \abh{$\texttt{cars}$, $\texttt{bicycles}$, $\texttt{trucks}$,} \textit{etc}.~\citep{liftsplat,CVT}.
    Some works also use \emph{instance} segmentation information to train their model where the centers of each distinct vehicle is known at train time \citep{fiery}.
    In all of our experiments, we place ourselves in the binary semantic segmentation setting of the meta-class \abh{$\texttt{vehicles}$}. This choice is made to have fair and consistent comparisons with our baselines \citep{liftsplat,CVT}, however, it should be noted that our model is not constrained to this setting.
    
    \item The levels of \textbf{visibility} of objects. Objects selected as ground truth, both for training and evaluating the model, differ in terms of their levels of visibility. Three options have been considered: objects that are in line-of-sight with the ego car's LiDAR~\citep{PON}, or objects with a nuScenes visibility above a defined threshold, either 0\%~\citep{liftsplat} or 40\%~\citep{CVT}. %
\end{itemize}

\updated{
\section{Extension to driveable area segmentation task}

\input{rebuttal/quali/quali}

In this section, we also provide results for the driveable area segmentation task, also addressed by CVT~\citep{CVT}. Contrary to vehicle segmentation, this task requires the network to do ``amodal completion'' to a high degree, i.e., to correctly estimate regions of the road despite parts of it being severely occluded. 

We followed the protocol of CVT~\citep{CVT} for this segmentation task; the ground truth is generated using HD-map's polygons from the dataset. We kept the same hyperparameters we used for the vehicle segmentation task, with a minor difference to the learning rate: we divide it by a factor 10 after 15 epochs (compared to a constant learning rate for vehicle segmentation).

Quantitative and qualitative results for this additional task are given respectively in \autoref{tab:driveable} and \autoref{fig:driveable}. When compared with CVT, we observe that LaRa achieves better performance (+0.9). Note that we do not do multi-tasking: following CVT~\citep{CVT}, we train a model specifically for the task of driveable area segmentation. The qualitative examples in \autoref{fig:driveable} are produced by fusing predictions from two models.

\begin{table}[!h]
\caption{\textbf{Driveable area segmentation}. Results (in IoU) on nuScenes.}
\smallskip
\label{tab:driveable}
\centering
\begin{tabular}{ll}
\toprule
Method & IoU  \\
\midrule
CVT    & 74.3 \\
LaRa (ours)    & \textbf{75.2} \\
\bottomrule
\end{tabular}
\end{table}

\section{A quantitative study of the influence of ray embedding on attention consistency across cameras}

\input{rebuttal/attention}

\begin{figure}[!h]
\centering
\includegraphics[width=0.8\linewidth]{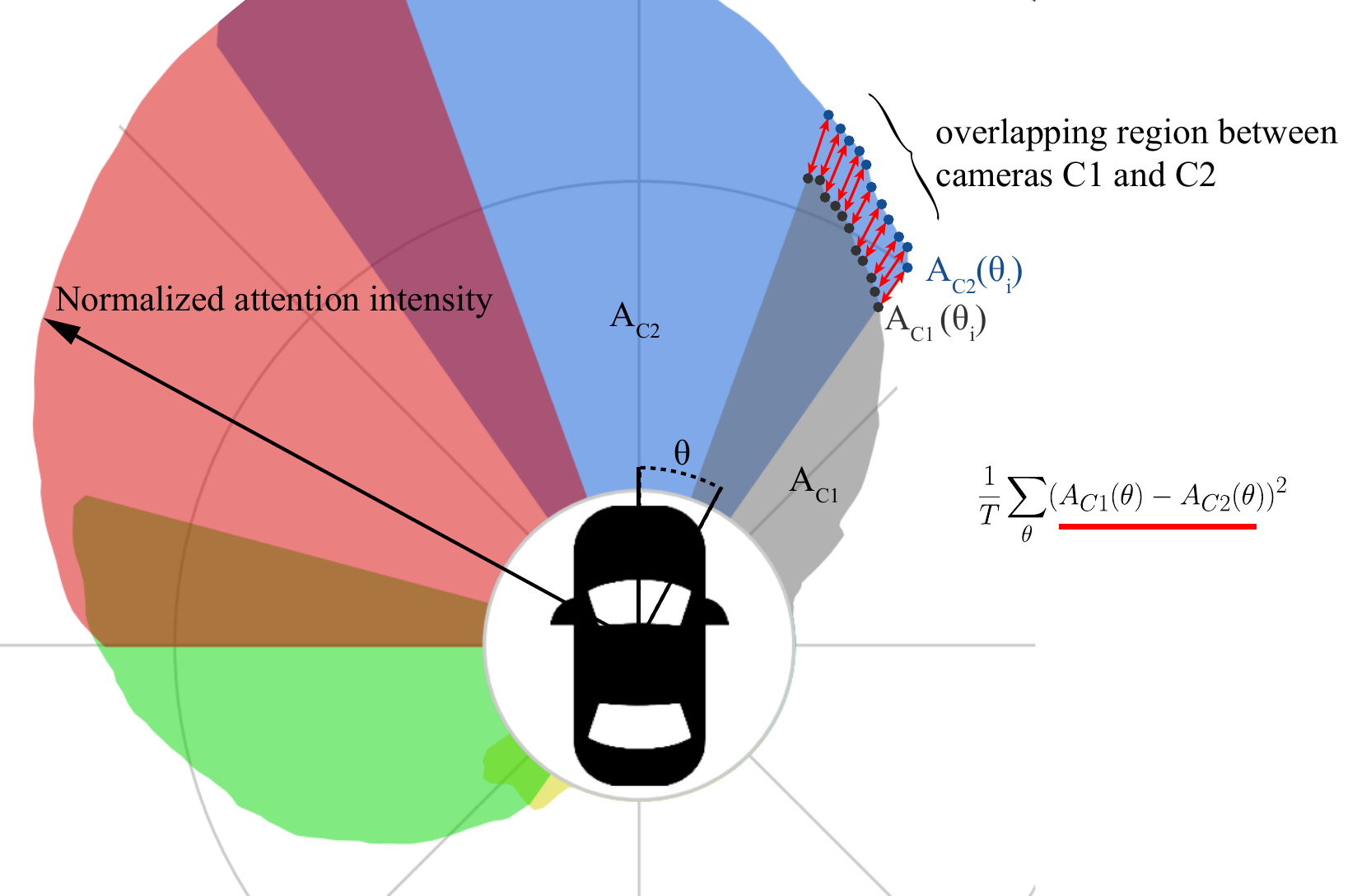}
\caption{\textbf{Measuring the attention consistency across cameras}. The proposed metric computes the Mean-Squared-Error (MSE) of the attention intensity on overlapping regions between cameras (as illustrated for two cameras and one latent and one attention head), and averages it over all cameras, latents, heads and scenes.}
\label{fig:mse_on_overlaps}
\end{figure}

In this section, we propose a quantitative analysis to support our claim that  ``our network is able to retrieve the pixel relationships between views thanks to our ray embedding'' (Sec. 4.3 in the main paper).

To this end, we introduce a metric that directly quantifies the consistency and alignment of attention values across camera by analyzing behavior in ``overlapping'' regions, i.e., regions seen by two different cameras. We provide a visual description of this metric and its computation in~\autoref{fig:mse_on_overlaps}.

In short, knowing the orientation of each camera, we compute the Mean Squared Error (MSE) of the directional attention intensity between cameras on their overlapping regions. This score is averaged for all the overlapping regions, latents and attention heads, and examples in the validation set. A score of zero indicates a perfect match of the attention levels on overlapping regions (i.e., across cameras). Results with this metric, reported in~\autoref{tab:mse_overlab}, show that our `Cam. rays' embedding is 10 times more ``consistent'' across cameras than the baseline `Fourier + Cam. idx'.

\begin{table}[!h]
\caption{
\textbf{Impact of ray embedding on cross-camera attention consistency.} Cross-camera attention consistency (measured with proposed MSE metric, see Fig.\,\ref{fig:attention_consistency}) on nuScene. 
}
\smallskip
\centering
\begin{tabular}{ll}
\toprule
Embedding & MSE on overlap  \\
\midrule
2D Fourier + Cam. idx    & 0.0896 \\
Cam. rays (ours)    & \textbf{0.0068} \\
\bottomrule
\end{tabular}
\label{tab:mse_overlab}
\end{table}

Additionally, we provide qualitative examples of the `Fourier + Cam. idx' embedding to compare against our ray embedding in \autoref{fig:attention_consistency}. Contrary to the attention yield by our ray embedding, the one derived from the `Fourier + Cam. idx' embedding is much more spread out and less consistent across cameras.

\section{Comparison to PETR encoding}

In PETR~\citep{PETR}, the embedding of each pixel is computed by sampling its ray given $D$  predefined depths. The 3D coordinates of the $D$ sampled points along the ray are normalized, concatenated, processed by an MLP and summed with the visual features. Conceptually, the embedding is a way to indicate to the network ``this pixel can observe these 3D points in the camera frustum space''.

The embedding in PETR differs in that it is limited by the sampling resolution (i.e., the $D$ predefined depths), as computation and memory footprint increase linearly with respect to $D$. In contrast, we showed that our constant-complexity embedding is effective as a 3D positional embedding.

In addition, we include quantitative results to compare PETR embedding against our ray embedding in~\autoref{tab:PETR}. We trained our model with PETR input embedding in place of ours. The results show that our ray embedding performs better (+72\%).

\begin{table}[!h]
\caption{\textbf{Impact of ray embedding on performance.} Vehicle segmentation performance (in IoU) for vehicle segmentation on nuScenes.}
\smallskip
\centering
\begin{tabular}{ll}
\toprule
Embedding & IoU  \\
\midrule
PETR~\citep{PETR}    & 34.8 \\
Cam. rays (ours)    & \textbf{35.4} \\
\bottomrule
\end{tabular}
\label{tab:PETR}
\end{table}
}

\section{Additional attention qualitative analysis}

We also provide additional analysis of attention maps for the multi-camera input shown in~\autoref{fig:attention_study:cam_band} with a network using 256 latents and 32 attention heads. As in the main paper, the polar plots represent the directional attention intensity, showing the directions the network attends the most. The contribution of each camera is indicated by a color code coherent with~\autoref{fig:attention_study:cam_band}. Each polar plot is oriented in an upward direction (i.e., the front of the car points upward).

\input{supp_figures/attention/attention}

\begin{figure}[h!]
    \centering
    \includegraphics[width=\textwidth]{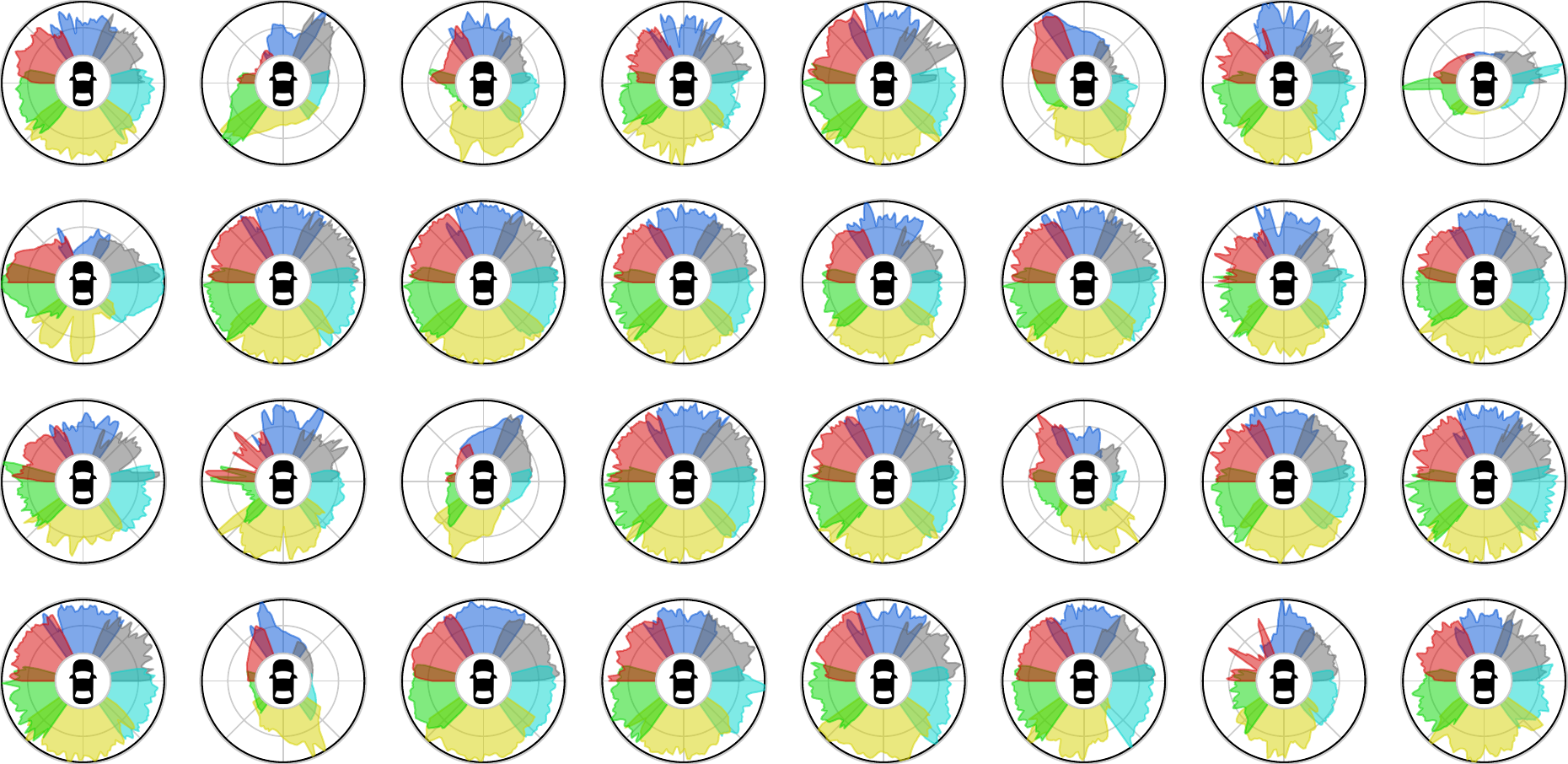}
    \caption{\textbf{Input-to-latent attention study --- average over latents.} These polar plots represent the directional attention intensity averaged over all the 256 latent vectors for each attention head. When averaging over latent vectors, we observe that each attention head generally covers all directions. This suggests that the latent vectors contain most of the directional information and that the whole scene is attended across the latent. More rarely, an attention head's polar plot will be directional but will maintain a level of generality by being symmetrical.}
    \label{fig:attention_study:allhead_avglatents}
\end{figure}

\begin{figure}[h!]
    \centering

    \includegraphics[width=\textwidth]{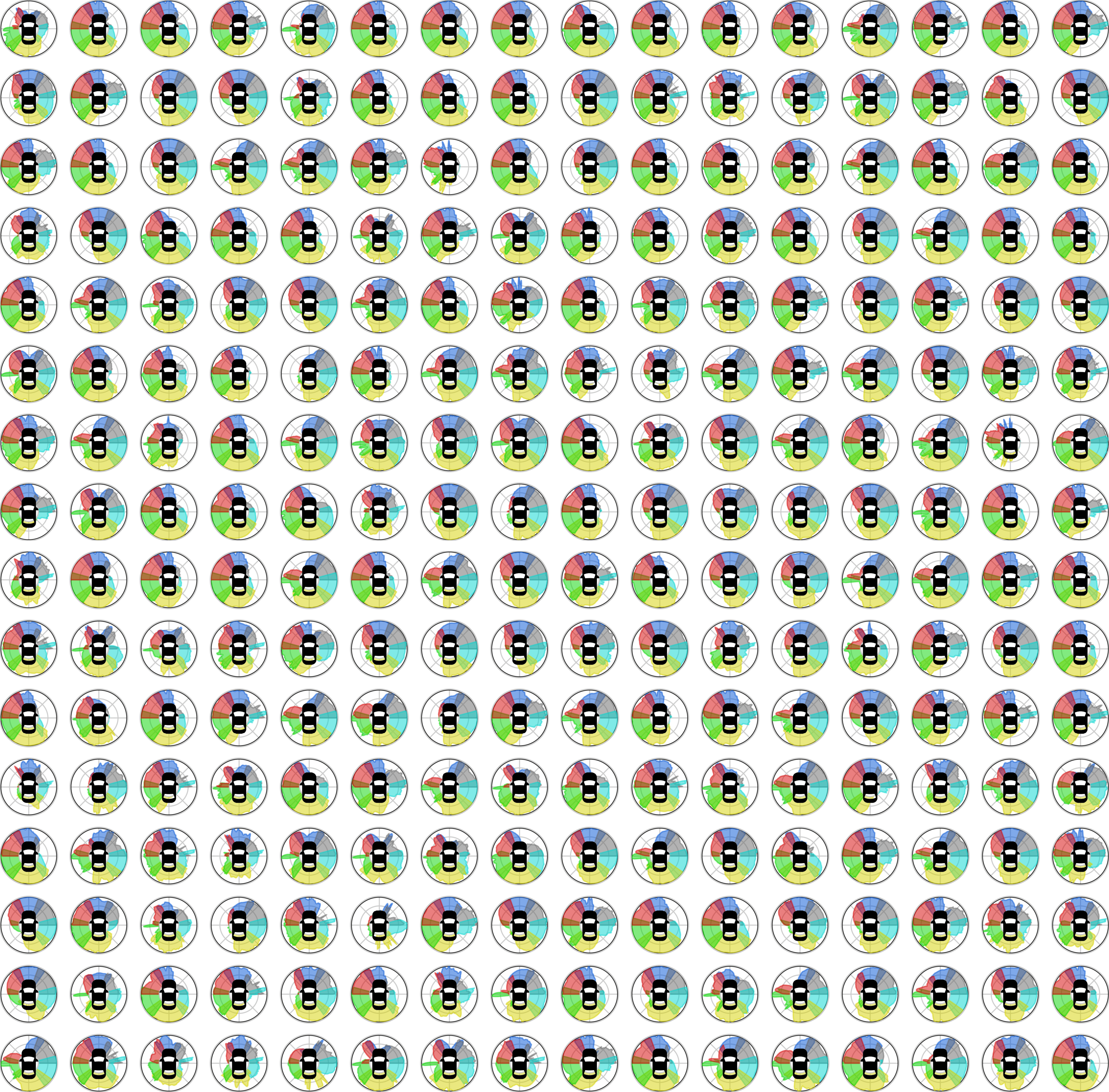}
    \caption{\textbf{Input-to-latent attention study --- average over heads.} These polar plots represent the directional attention intensity averaged over all attention heads for the 32 attention heads. When averaging over attention heads, we observe that the average attention spans over half of the scene. This allows latent vectors to extract long-range context between views with the capacity to disambiguate them.}
    \label{fig:attention_study:avghead_alllatents}
\end{figure}

\begin{figure}[h!]
    \centering

    \includegraphics[width=\textwidth]{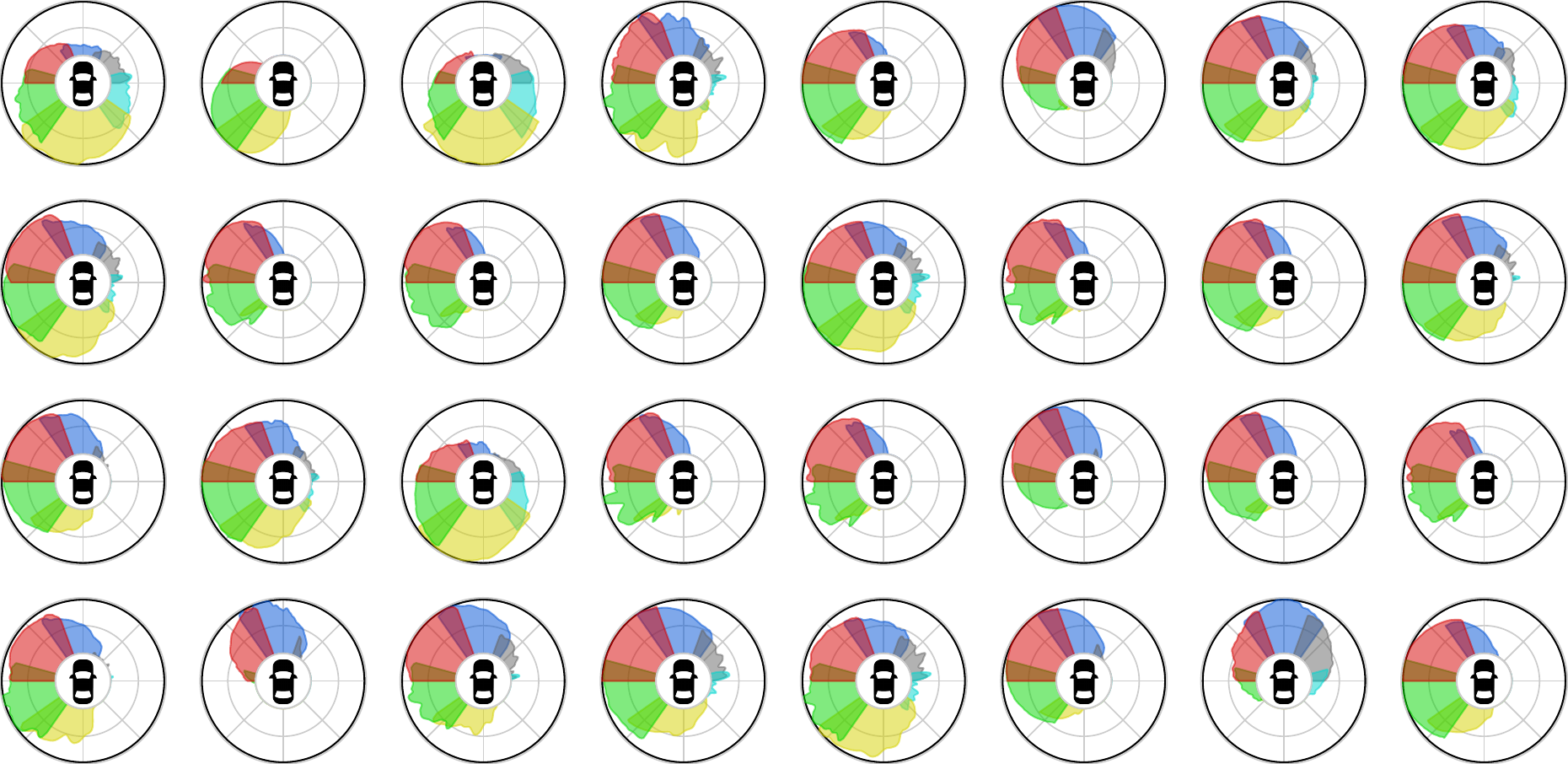}
    \caption{\textbf{Input-to-latent attention study --- all the attention heads of a latent vector.} These polar plots represent the directional attention intensity of the 32 attention heads for a randomly chosen latent vector (latent vector \#10). As shown in \autoref{fig:attention_study:avghead_alllatents}, one latent vector approximately covers half of the scene over its attention heads.}
    \label{fig:attention_study:allhead_l10}
\end{figure}

\begin{figure}[h!]
    \centering
    \includegraphics[width=\textwidth]{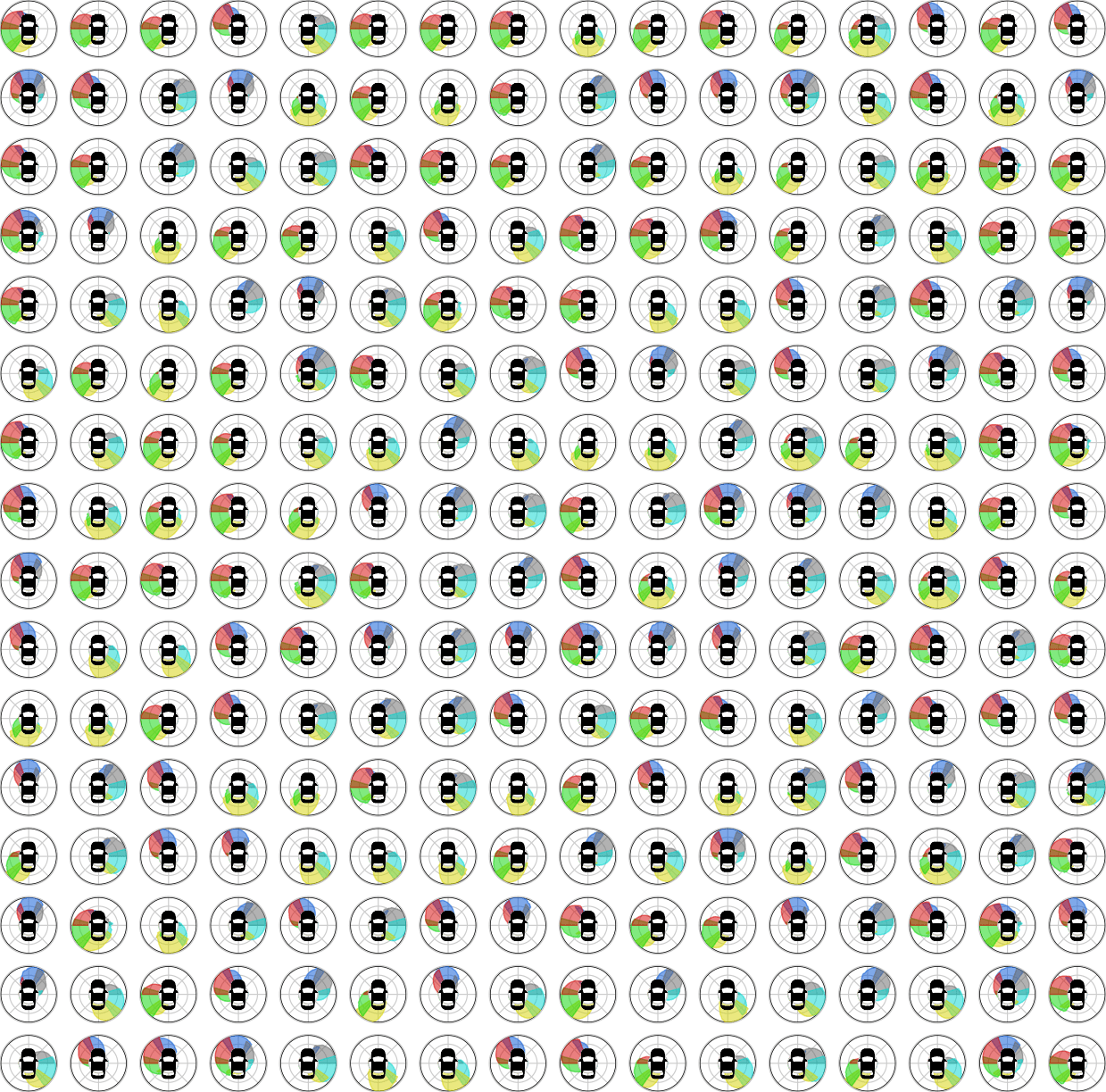}
    \caption{\textbf{Input-to-latent attention study --- all the latent vectors for an attention head.} These polar plots represent the directional attention intensity of the 256 latent vectors for a randomly chosen attention head (head \#4). As shown in \autoref{fig:attention_study:allhead_avglatents}, one attention head generally covers the full scene over the latent vectors.}
    \label{fig:attention_study:h4_alllatents}
\end{figure}

\section{Additional qualitative examples}

We also provide videos of our segmentation results on complex scenes in various visual conditions (daylight, rain, night). In these videos, we compare against our two baselines CVT~\citep{CVT} and Lift-Splat~\citep{liftsplat}. For a fair comparison, we use our model trained with visibility $> 40\%$ against CVT and $> 0\%$ against Lift-Splat.

%% file: rebuttal/quali/quali.tex
\newcommand{\BEVimage}[1]{\includegraphics[width=0.18\linewidth]{#1}}

\definecolor{colorF}{HTML}{0053D6}
\definecolor{colorFL}{HTML}{D60000}
\definecolor{colorFR}{HTML}{666666}
\definecolor{colorB}{HTML}{D6D200}
\definecolor{colorBL}{HTML}{04D600}
\definecolor{colorBR}{HTML}{00D6CF}

\begin{figure*}[!h]
\vspace{-5pt}
\centering
\resizebox{\textwidth}{!}{
\begin{tikzpicture}[
    every node/.style={inner sep=0,outer sep=2},
    tag/.style = {
        fill=black,
        inner sep=1pt, 
        outer sep=0pt,
        yshift=2pt,
        xshift=2pt,
        font=\scriptsize,
        text=white,
        align=center,
        anchor=south,
    },
    camtag/.style = {
        inner sep=1pt, 
        outer sep=0pt,
        font=\tiny,
        text=white,
        align=center,
        fill opacity=0.5,
        text opacity=1
    },
    bev/.style = {
        anchor=west,
        rectangle,
        draw=black,
        line width=1pt
    },
]

    \node (CamSeq2Idx20) {\camsimage{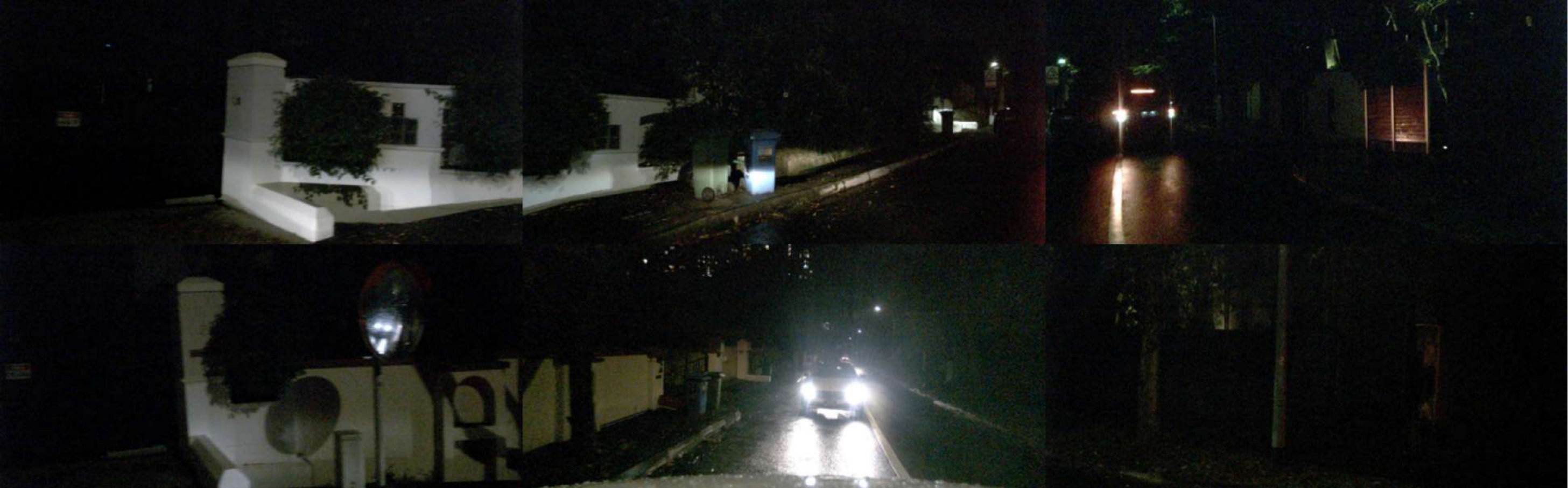}};
    \node[rectangle, camtag, fill=colorFL,  anchor=north west, yshift=-2pt, xshift=20pt] (CamSeq2Idx20_FL) at (CamSeq2Idx20.north west) {front left};
    \node[rectangle, camtag, fill=colorF, anchor=north, yshift=-2pt] (CamSeq2Idx20_F) at (CamSeq2Idx20.north) {front};
    \node[rectangle, camtag, fill=colorFR, anchor=north east, yshift=-2pt, xshift=-20pt] (CamSeq2Idx20_FR) at (CamSeq2Idx20.north east) {front right};
    \node[rectangle, camtag, fill=colorBL, anchor=west, yshift=-3pt, xshift=20pt] (CamSeq2Idx20_BL) at (CamSeq2Idx20.west) {back left};
    \node[rectangle, camtag, fill=colorB, anchor=center, yshift=-3pt] (CamSeq2Idx20_B) at (CamSeq2Idx20.center) {back};
    \node[rectangle, camtag, fill=colorBR, anchor=east, yshift=-3pt, xshift=-20pt] (CamSeq2Idx20_BR) at (CamSeq2Idx20.east) {back right};

    \node [bev] (GTSeq2Idx20) at (CamSeq2Idx20.east) {\BEVimage{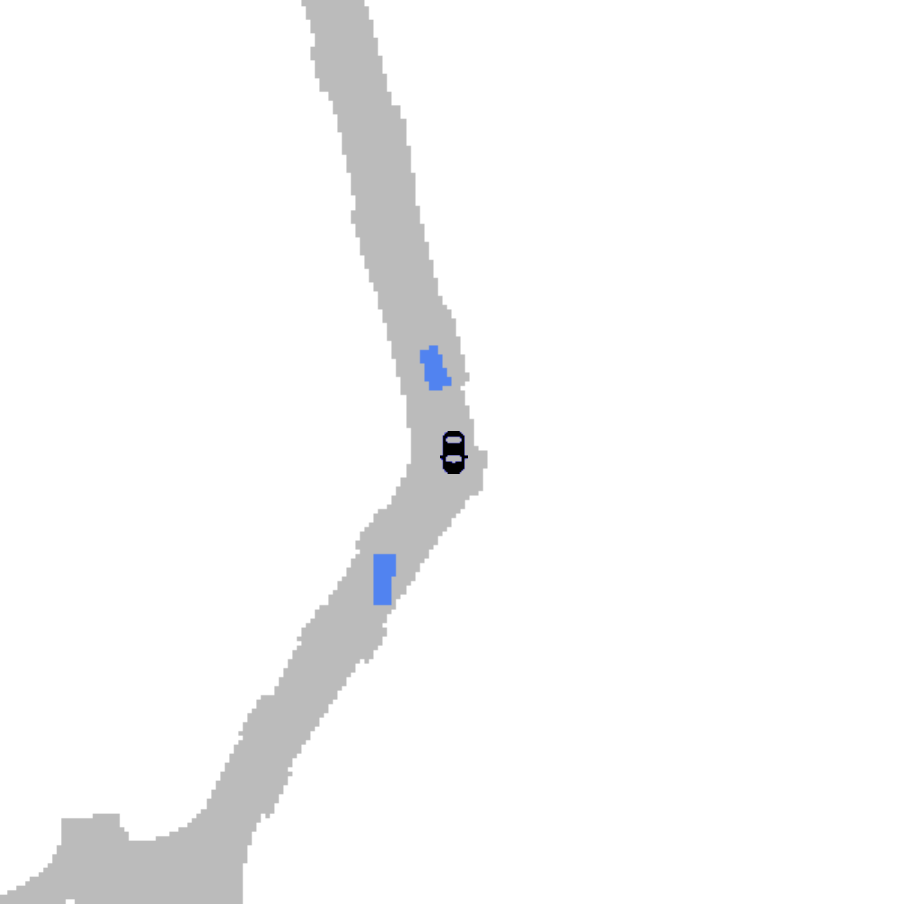}};
    \node[rectangle, tag] (GTSeq2Idx20_tag) at (GTSeq2Idx20.south) { GT };

    \node [bev] (PredSeq2Idx20) at (GTSeq2Idx20.east) {\BEVimage{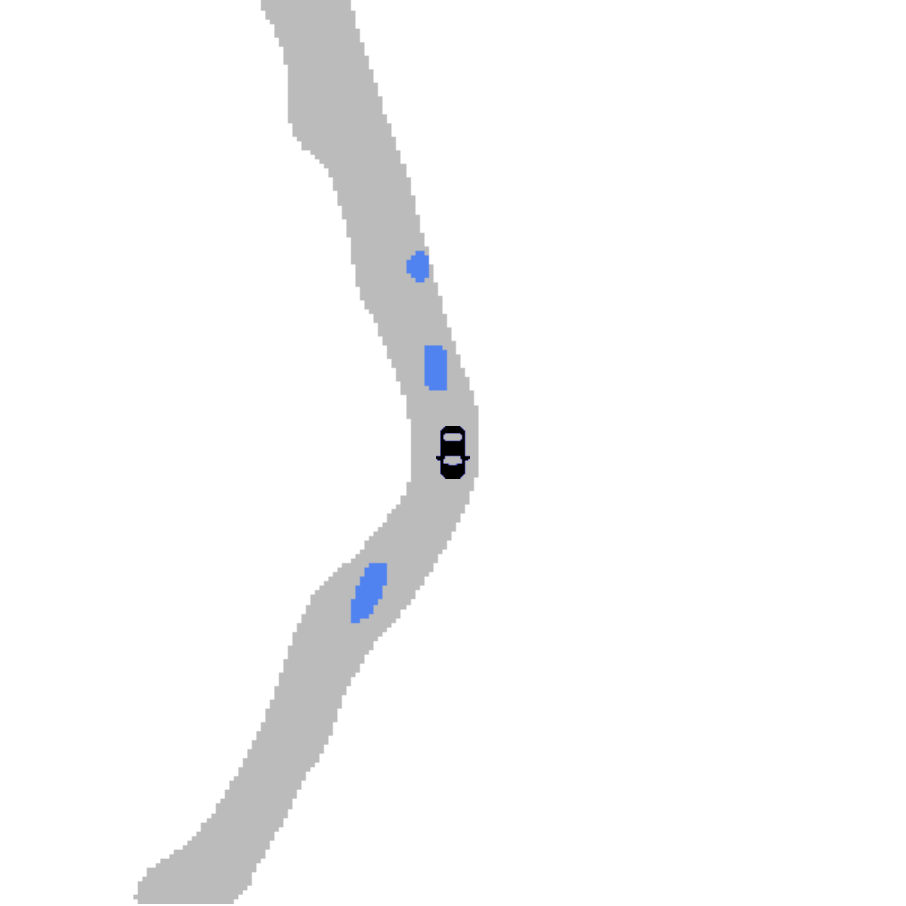}};
    \node[rectangle, tag] (PredSeq2Idx20_tag) at (PredSeq2Idx20.south) { Pred };

    \node [below=0.6cm of CamSeq2Idx20](CamSeq5Idx25) {\camsimage{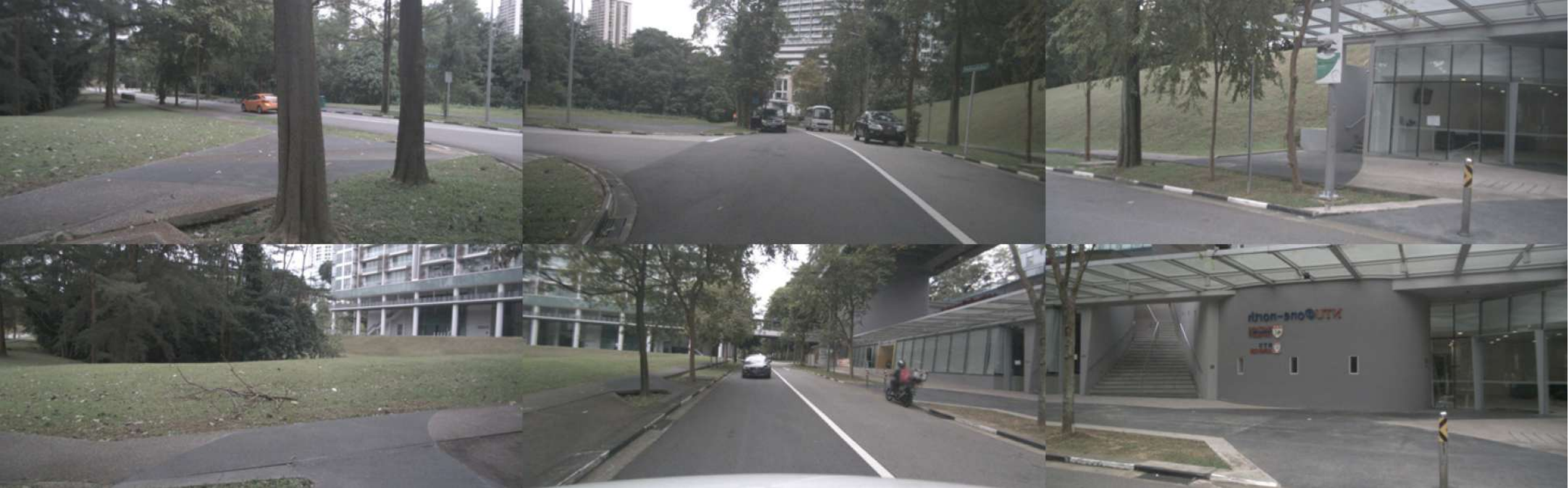}};
    \node[rectangle, camtag, fill=colorFL,  anchor=north west, yshift=-2pt, xshift=20pt] (CamSeq5Idx25_FL) at (CamSeq5Idx25.north west) {front left};
    \node[rectangle, camtag, fill=colorF, anchor=north, yshift=-2pt] (CamSeq5Idx25_F) at (CamSeq5Idx25.north) {front};
    \node[rectangle, camtag, fill=colorFR, anchor=north east, yshift=-2pt, xshift=-20pt] (CamSeq5Idx25_FR) at (CamSeq5Idx25.north east) {front right};
    \node[rectangle, camtag, fill=colorBL, anchor=west, yshift=-3pt, xshift=20pt] (CamSeq5Idx25_BL) at (CamSeq5Idx25.west) {back left};
    \node[rectangle, camtag, fill=colorB, anchor=center, yshift=-3pt] (CamSeq5Idx25_B) at (CamSeq5Idx25.center) {back};
    \node[rectangle, camtag, fill=colorBR, anchor=east, yshift=-3pt, xshift=-20pt] (CamSeq5Idx25_BR) at (CamSeq5Idx25.east) {back right};

    \node [bev] (GTSeq5Idx25) at (CamSeq5Idx25.east) {\BEVimage{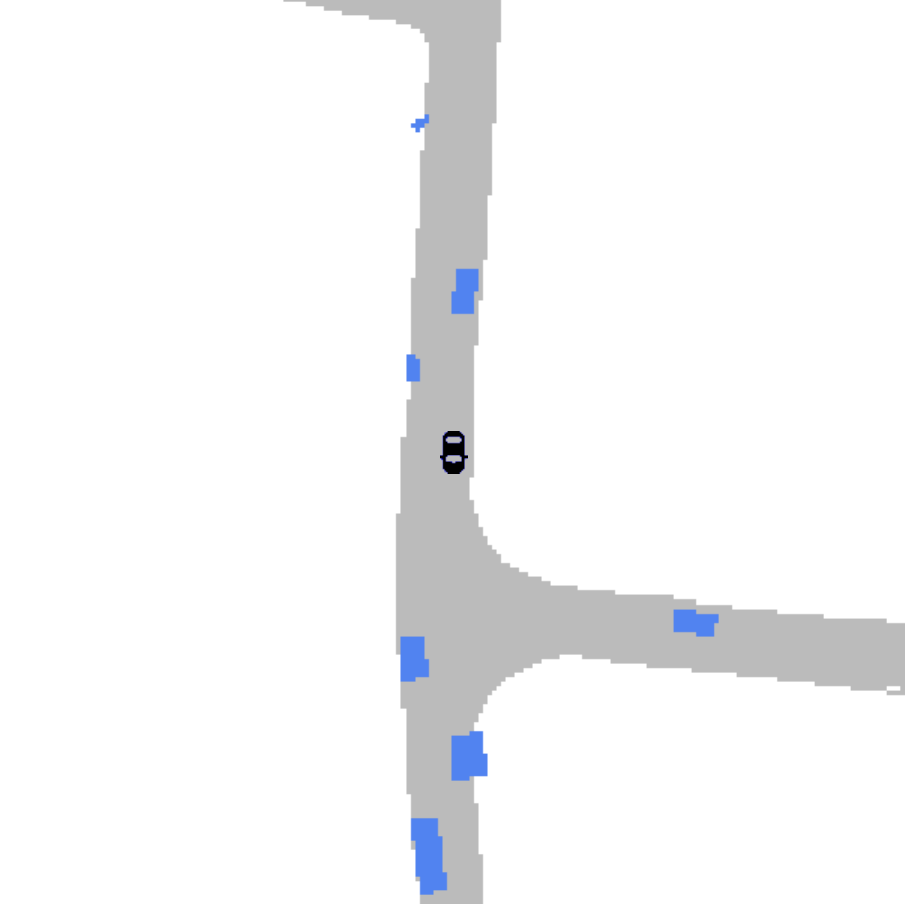}};
    \node[rectangle, tag] (GTSeq5Idx25_tag) at (GTSeq5Idx25.south) { GT };

    \node [bev] (PredSeq5Idx25) at (GTSeq5Idx25.east) {\BEVimage{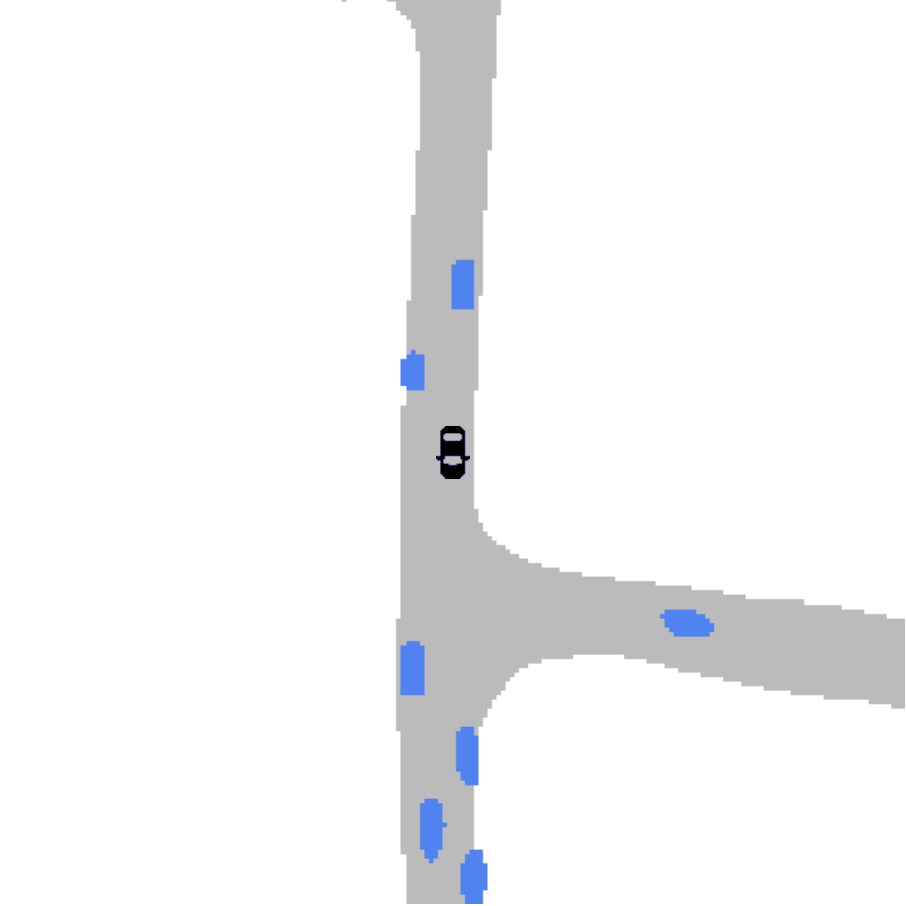}};
    \node[rectangle, tag] (PredSeq5Idx25_tag) at (PredSeq5Idx25.south) { Pred };

    \node [below=0.6cm of CamSeq5Idx25](CamSeq25Idx5) {\camsimage{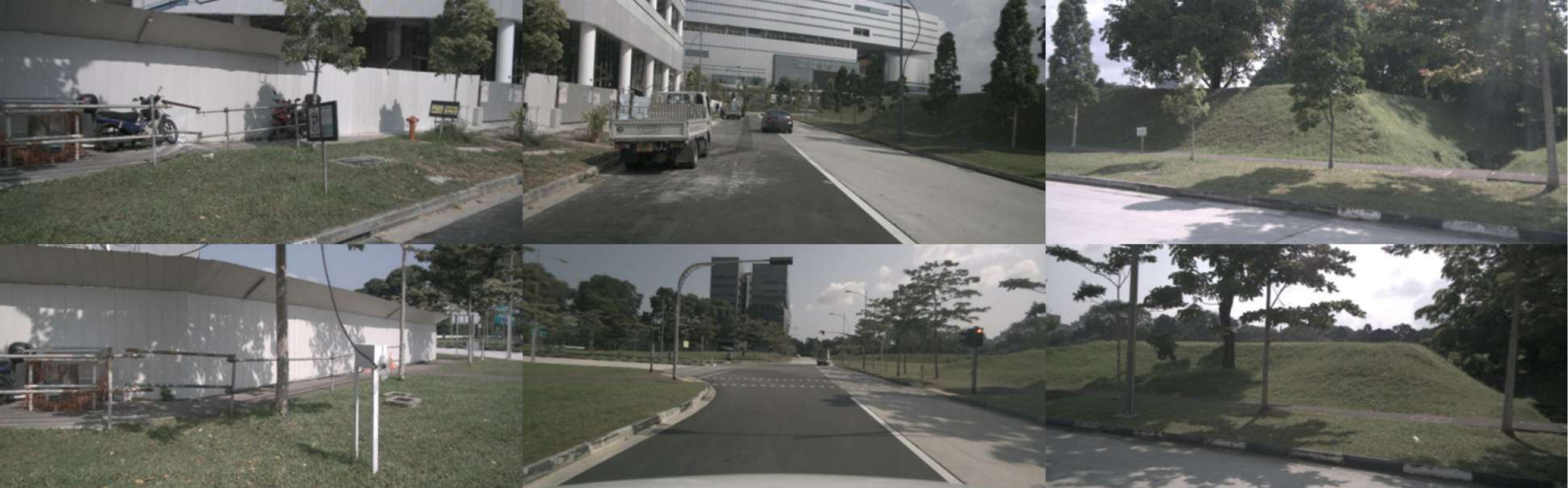}};
    \node[rectangle, camtag, fill=colorFL,  anchor=north west, yshift=-2pt, xshift=20pt] (CamSeq25Idx5_FL) at (CamSeq25Idx5.north west) {front left};
    \node[rectangle, camtag, fill=colorF, anchor=north, yshift=-2pt] (CamSeq25Idx5_F) at (CamSeq25Idx5.north) {front};
    \node[rectangle, camtag, fill=colorFR, anchor=north east, yshift=-2pt, xshift=-20pt] (CamSeq25Idx5_FR) at (CamSeq25Idx5.north east) {front right};
    \node[rectangle, camtag, fill=colorBL, anchor=west, yshift=-3pt, xshift=20pt] (CamSeq25Idx5_BL) at (CamSeq25Idx5.west) {back left};
    \node[rectangle, camtag, fill=colorB, anchor=center, yshift=-3pt] (CamSeq25Idx5_B) at (CamSeq25Idx5.center) {back};
    \node[rectangle, camtag, fill=colorBR, anchor=east, yshift=-3pt, xshift=-20pt] (CamSeq25Idx5_BR) at (CamSeq25Idx5.east) {back right};

    \node [bev] (GTSeq25Idx5) at (CamSeq25Idx5.east) {\BEVimage{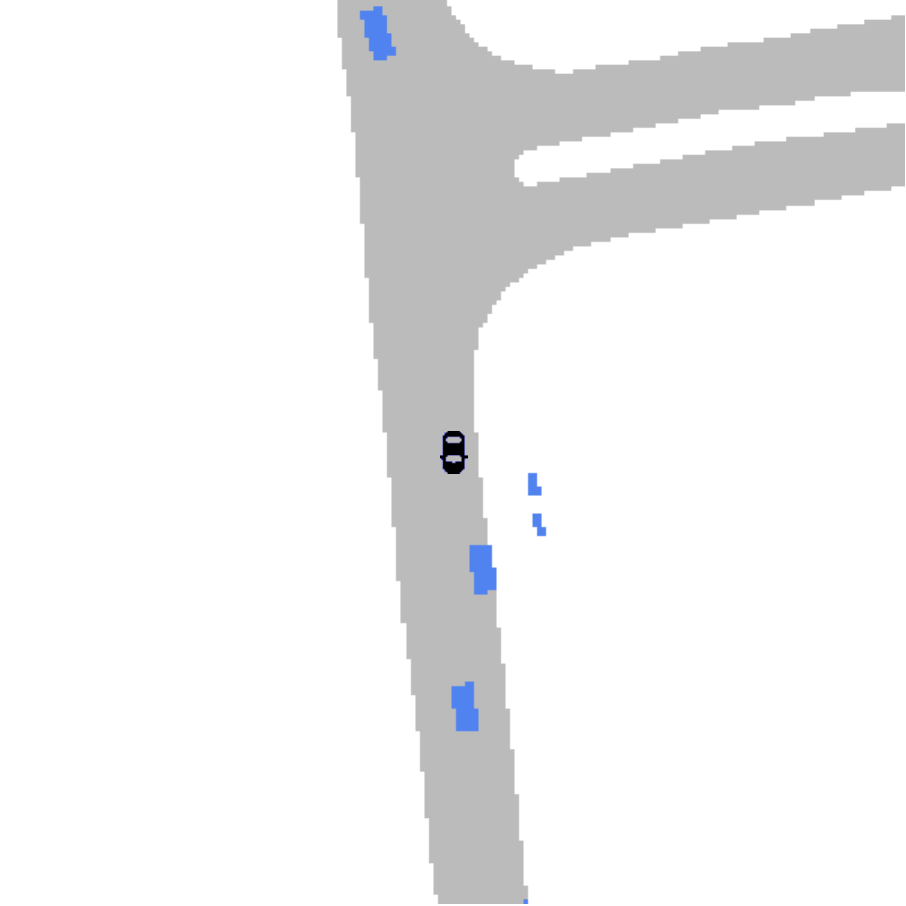}};
    \node[rectangle, tag] (GTSeq25Idx5_tag) at (GTSeq25Idx5.south) { GT };

    \node [bev] (PredSeq25Idx5) at (GTSeq25Idx5.east) {\BEVimage{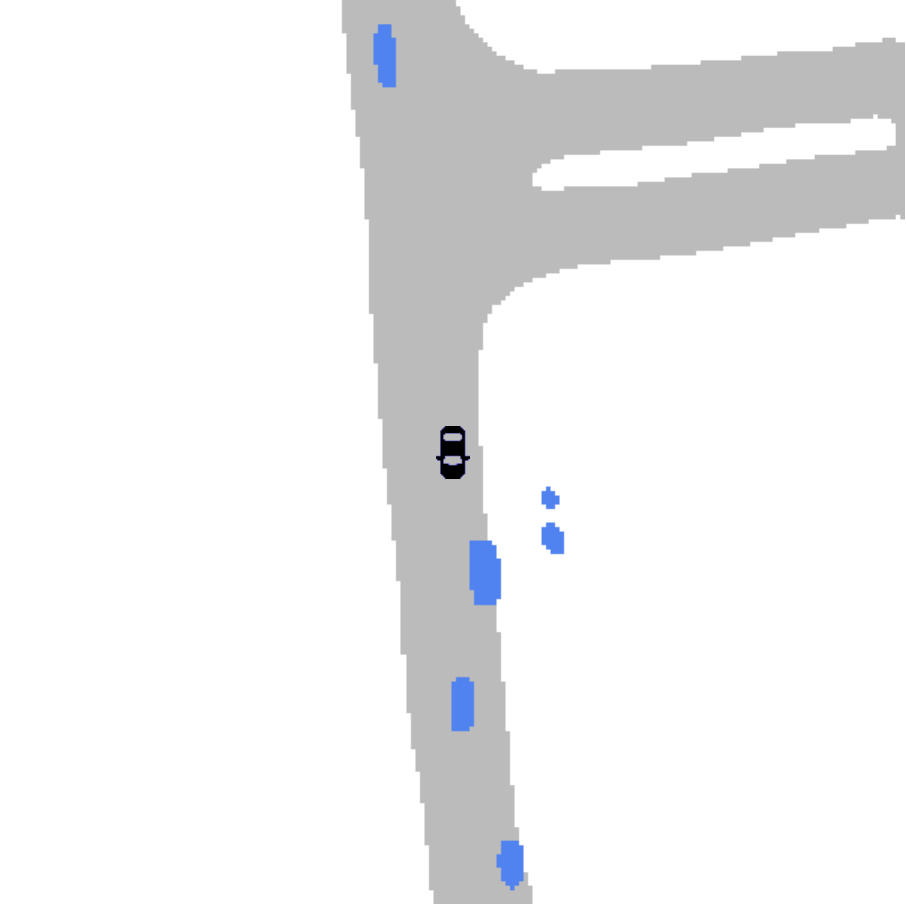}};
    \node[rectangle, tag] (PredSeq25Idx5_tag) at (PredSeq25Idx5.south) { Pred };
    
    \node [below=0.6cm of CamSeq25Idx5](CamSeq26Idx5) {\camsimage{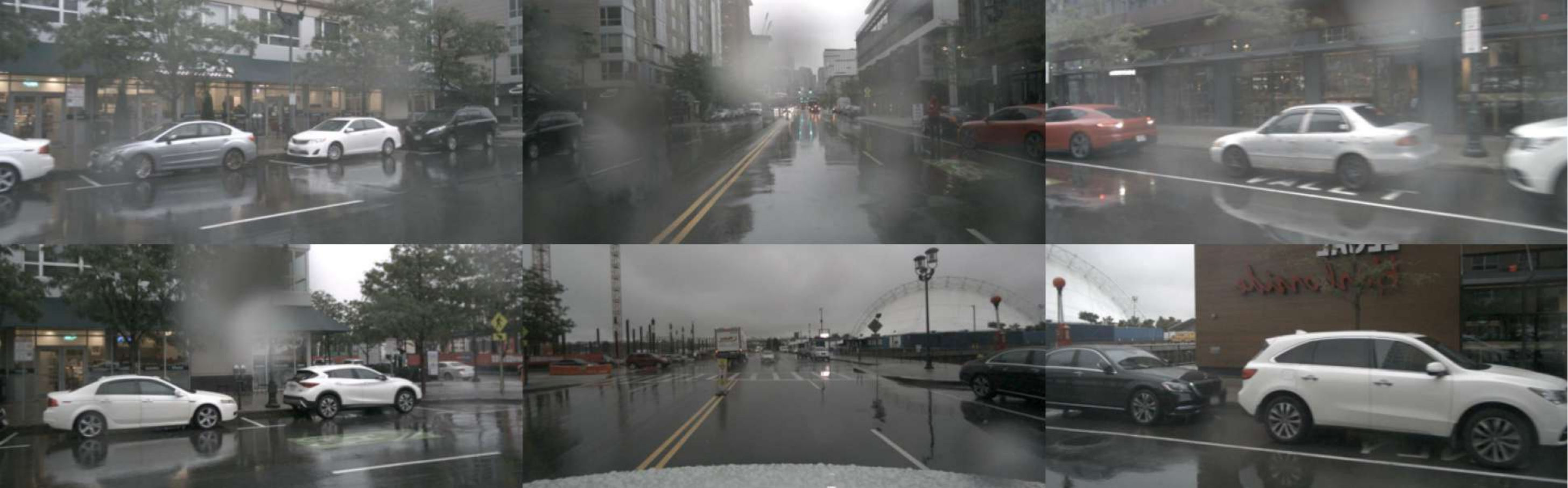}};
    \node[rectangle, camtag, fill=colorFL,  anchor=north west, yshift=-2pt, xshift=20pt] (CamSeq26Idx5_FL) at (CamSeq26Idx5.north west) {front left};
    \node[rectangle, camtag, fill=colorF, anchor=north, yshift=-2pt] (CamSeq26Idx5_F) at (CamSeq26Idx5.north) {front};
    \node[rectangle, camtag, fill=colorFR, anchor=north east, yshift=-2pt, xshift=-20pt] (CamSeq26Idx5_FR) at (CamSeq26Idx5.north east) {front right};
    \node[rectangle, camtag, fill=colorBL, anchor=west, yshift=-3pt, xshift=20pt] (CamSeq26Idx5_BL) at (CamSeq26Idx5.west) {back left};
    \node[rectangle, camtag, fill=colorB, anchor=center, yshift=-3pt] (CamSeq26Idx5_B) at (CamSeq26Idx5.center) {back};
    \node[rectangle, camtag, fill=colorBR, anchor=east, yshift=-3pt, xshift=-20pt] (CamSeq26Idx5_BR) at (CamSeq26Idx5.east) {back right};

    \node [bev] (GTSeq26Idx5) at (CamSeq26Idx5.east) {\BEVimage{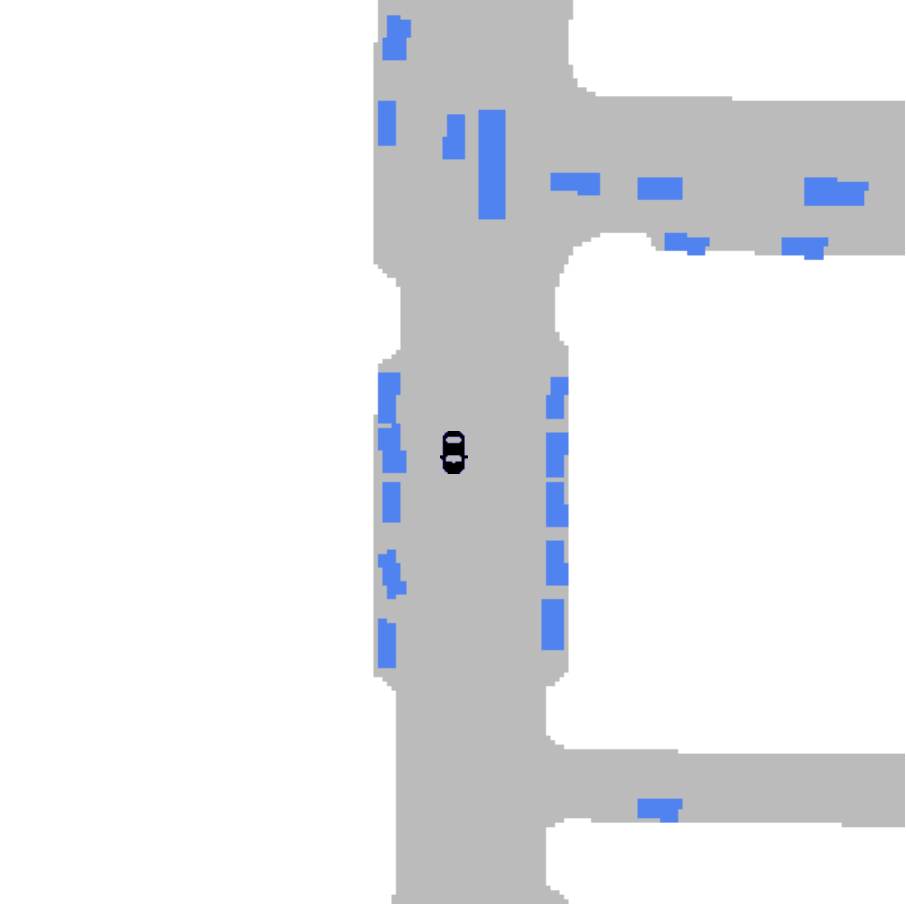}};
    \node[rectangle, tag] (GTSeq26Idx5_tag) at (GTSeq26Idx5.south) { GT };

    \node [bev] (PredSeq26Idx5) at (GTSeq26Idx5.east) {\BEVimage{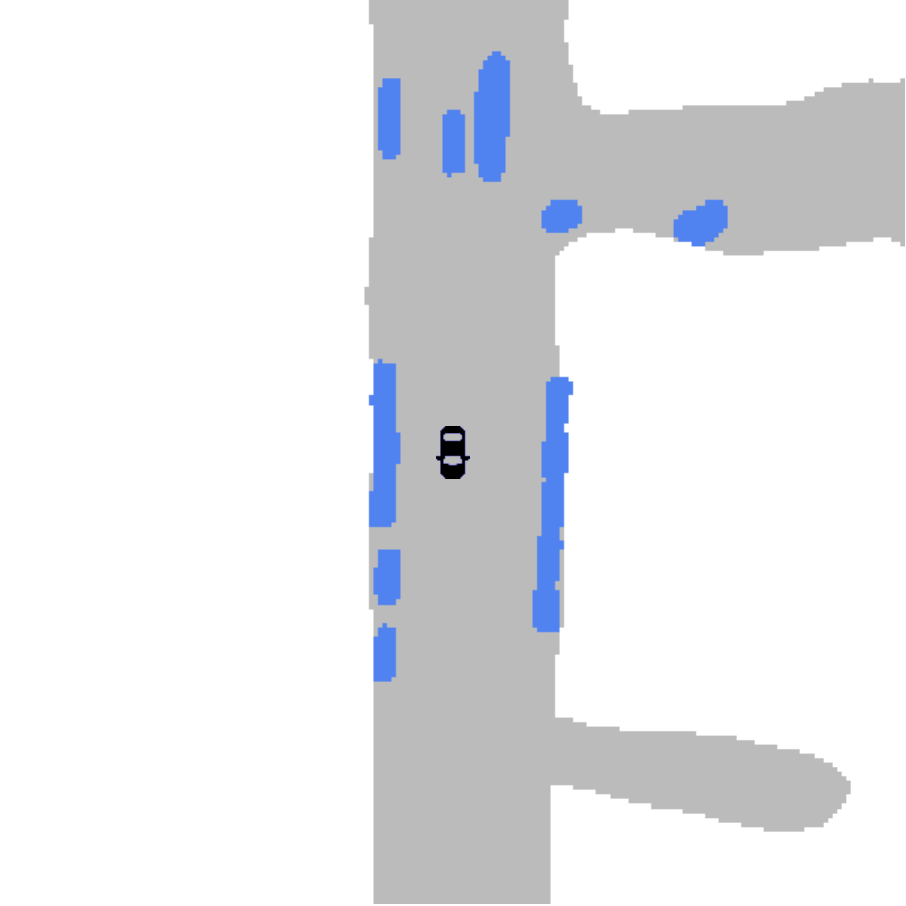}};
    \node[rectangle, tag] (PredSeq26Idx5_tag) at (PredSeq26Idx5.south) { Pred };

\end{tikzpicture}
}
\caption{\textbf{Qualitative results on complex scenes.} We show the six camera views surrounding the vehicle along with segmentation ground truth for reference. Vehicles are shown in blue and driveable area in gray. Vehicles and driveable area predictions are from two different models trained independently for their respective ground-truth, the predictions are then merged for vizualization purpose.
The ego vehicle is located in the center and facing downwards. Predictions of both driveable area and vehicle segmentation are thresholded at 0.5 for visualization purpose.}
\label{fig:driveable}
\end{figure*}

%% file: rebuttal/attention.tex
\definecolor{colorF}{HTML}{0053D6}
\definecolor{colorFL}{HTML}{D60000}
\definecolor{colorFR}{HTML}{666666}
\definecolor{colorB}{HTML}{D6D200}
\definecolor{colorBL}{HTML}{04D600}
\definecolor{colorBR}{HTML}{00D6CF}

\begin{figure*}[!h]
\vspace{-5pt}
\centering
\resizebox{\textwidth}{!}{
\begin{tikzpicture}[
    every node/.style={inner sep=0,outer sep=2},
    label/.style = {
        inner sep=2pt,
        font=\scriptsize,
        align=center,
    },
    scores/.style = {
        inner sep=2pt, 
        outer sep=0pt,
        font=\footnotesize,
        text=white,
        align=center,
        anchor=north,
    },
    tag/.style = {
        fill=black,
        inner sep=1pt, 
        outer sep=0pt,
        yshift=2pt,
        xshift=2pt,
        font=\scriptsize,
        text=white,
        align=center,
        anchor=south west,
    },
    camtag/.style = {
        inner sep=1pt, 
        outer sep=0pt,
        font=\tiny,
        text=white,
        align=center,
        fill opacity=0.5,
        text opacity=1
    },
]

    \node (Cam10) {\camsimage{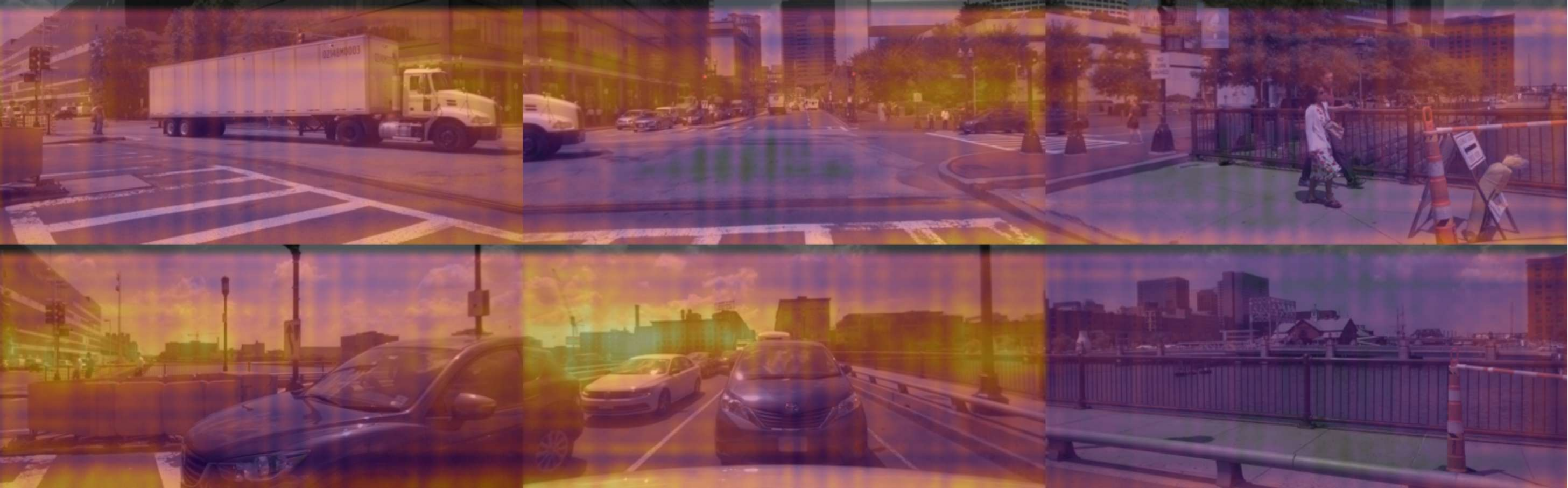}};
    \node[rectangle, tag] (Cam10_tag) at (Cam10.south west) {n:80 h:10};
    \node[rectangle, camtag, fill=colorFL,  anchor=north west, yshift=-2pt, xshift=20pt] (Cam10_FL) at (Cam10.north west) {front left};
    \node[rectangle, camtag, fill=colorF, anchor=north, yshift=-2pt] (Cam10_F) at (Cam10.north) {front};
    \node[rectangle, camtag, fill=colorFR, anchor=north east, yshift=-2pt, xshift=-20pt] (Cam10_FR) at (Cam10.north east) {front right};
    \node[rectangle, camtag, fill=colorBL, anchor=west, yshift=-3pt, xshift=20pt] (Cam10_BL) at (Cam10.west) {back left};
    \node[rectangle, camtag, fill=colorB, anchor=center, yshift=-3pt] (Cam10_B) at (Cam10.center) {back};
    \node[rectangle, camtag, fill=colorBR, anchor=east, yshift=-3pt, xshift=-20pt] (Cam10_BR) at (Cam10.east) {back right};
    
    \node [anchor=west] (Polarl10h5) at (Cam10.east) {\polarimage{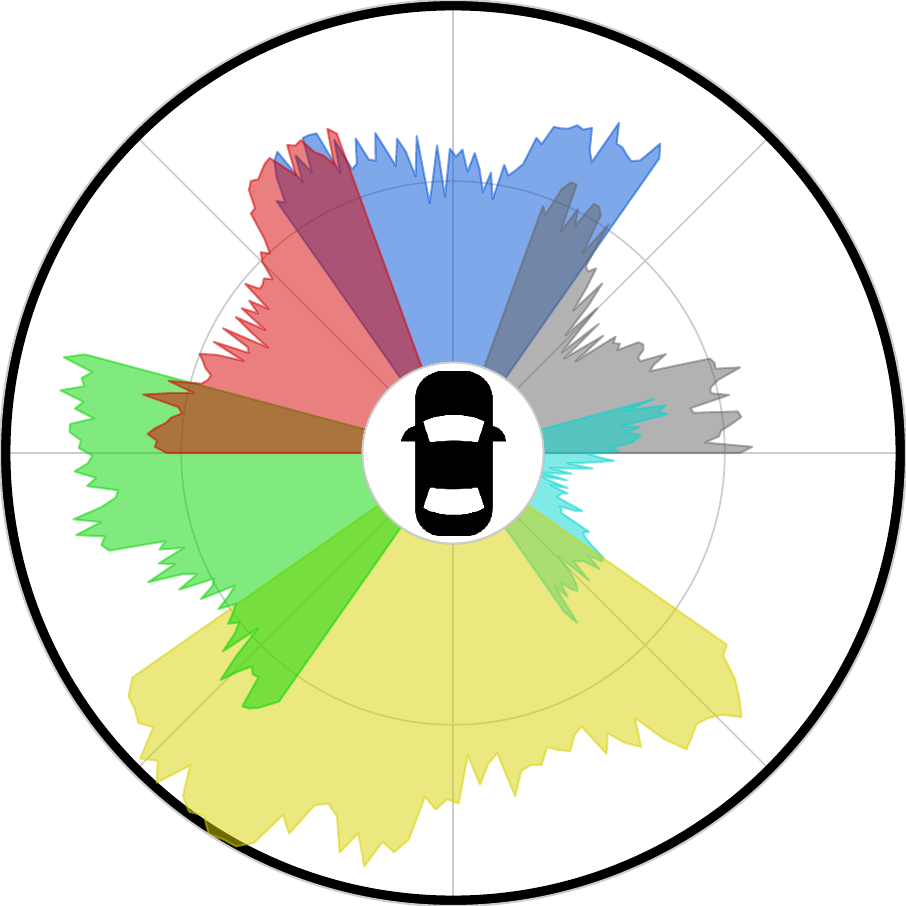}};
    \node[rectangle, tag] (Polarl10h5_tag) at (Polarl10h5.south west) {n:80 h:10};

    \node [anchor=north] (Cam50) at (Cam10.south) {\camsimage{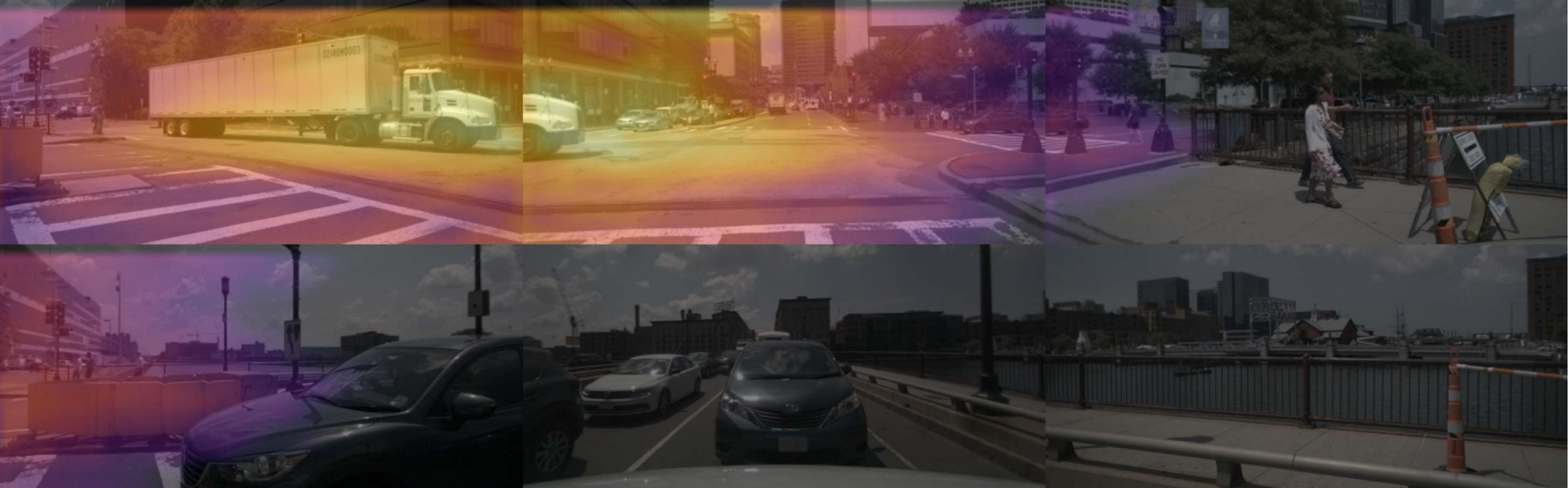}};
    \node[rectangle, tag] (Cam50_tag) at (Cam50.south west) {n:10 h:5};
    \node[rectangle, camtag, fill=colorFL,  anchor=north west, yshift=-2pt, xshift=20pt] (Cam50_FL) at (Cam50.north west) {front left};
    \node[rectangle, camtag, fill=colorF, anchor=north, yshift=-2pt] (Cam50_F) at (Cam50.north) {front};
    \node[rectangle, camtag, fill=colorFR, anchor=north east, yshift=-2pt, xshift=-20pt] (Cam50_FR) at (Cam50.north east) {front right};
    \node[rectangle, camtag, fill=colorBL, anchor=west, yshift=-3pt, xshift=20pt] (Cam50_BL) at (Cam50.west) {back left};
    \node[rectangle, camtag, fill=colorB, anchor=center, yshift=-3pt] (Cam50_B) at (Cam50.center) {back};
    \node[rectangle, camtag, fill=colorBR, anchor=east, yshift=-3pt, xshift=-20pt] (Cam50_BR) at (Cam50.east) {back right};
    
    \node [anchor=west] (Polarl50h30) at (Cam50.east) {\polarimage{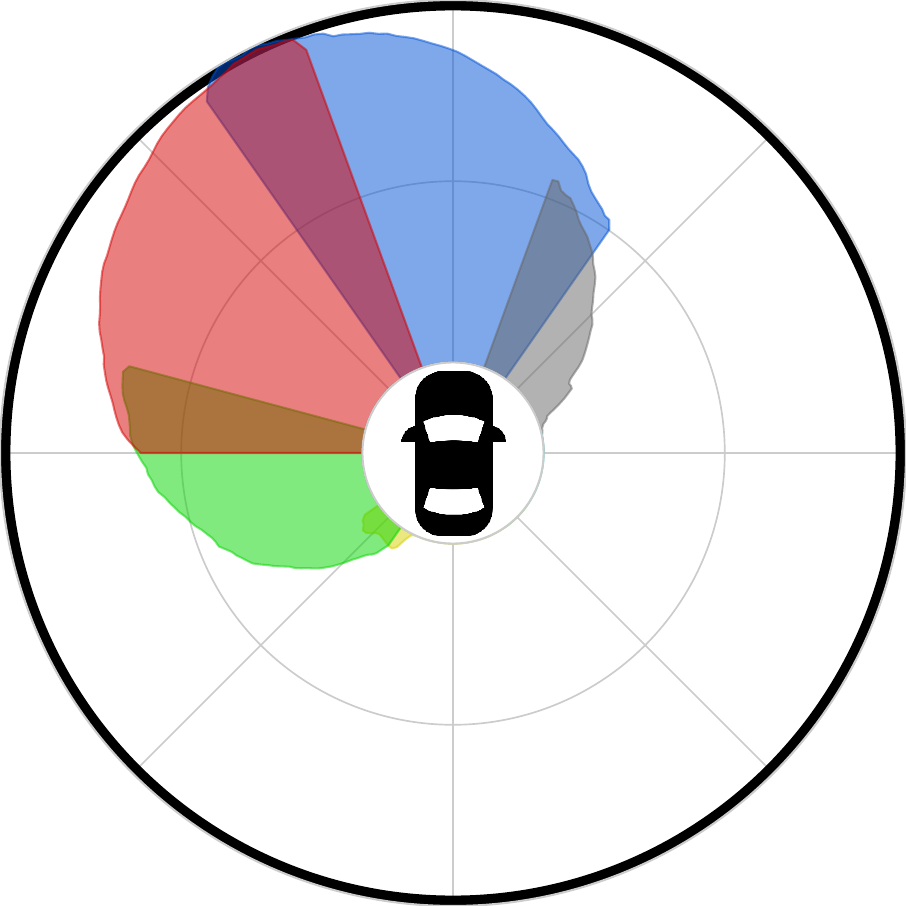}};
    \node[rectangle, tag] (Polarl50h30_tag) at (Polarl50h30.south west) {n:10 h:5};

\end{tikzpicture}
}
\caption{\textbf{Input-to-latent attention study.} Analysis of attention maps for two networks trained with different input embeddings. Top row is with `Fourier + Cam.\ idx' and bottom row is with our proposed `Cam.\ rays' embedding. The attention for one attention head and one latent is shown on the left superimposed with RGB images. The polar plots represent the directional attention intensity for one attention head with one latent vector. 
The radial distance is proportional to the attention level and shows the directions the network attends to the most.}
\label{fig:attention_consistency}
\end{figure*}

%% file: supp_figures/attention/attention.tex
\definecolor{colorF}{HTML}{0053D6}
\definecolor{colorFL}{HTML}{D60000}
\definecolor{colorFR}{HTML}{666666}
\definecolor{colorB}{HTML}{D6D200}
\definecolor{colorBL}{HTML}{04D600}
\definecolor{colorBR}{HTML}{00D6CF}

\begin{figure*}[b]
\centering

\begin{tikzpicture}[
    every node/.style={inner sep=0,outer sep=2},
    camtag/.style = {
        inner sep=1pt, 
        outer sep=0pt,
        text=white,
        align=center,
        fill opacity=0.5,
        text opacity=1
    },
]

    \node (Cam) {\includegraphics[width=\linewidth]{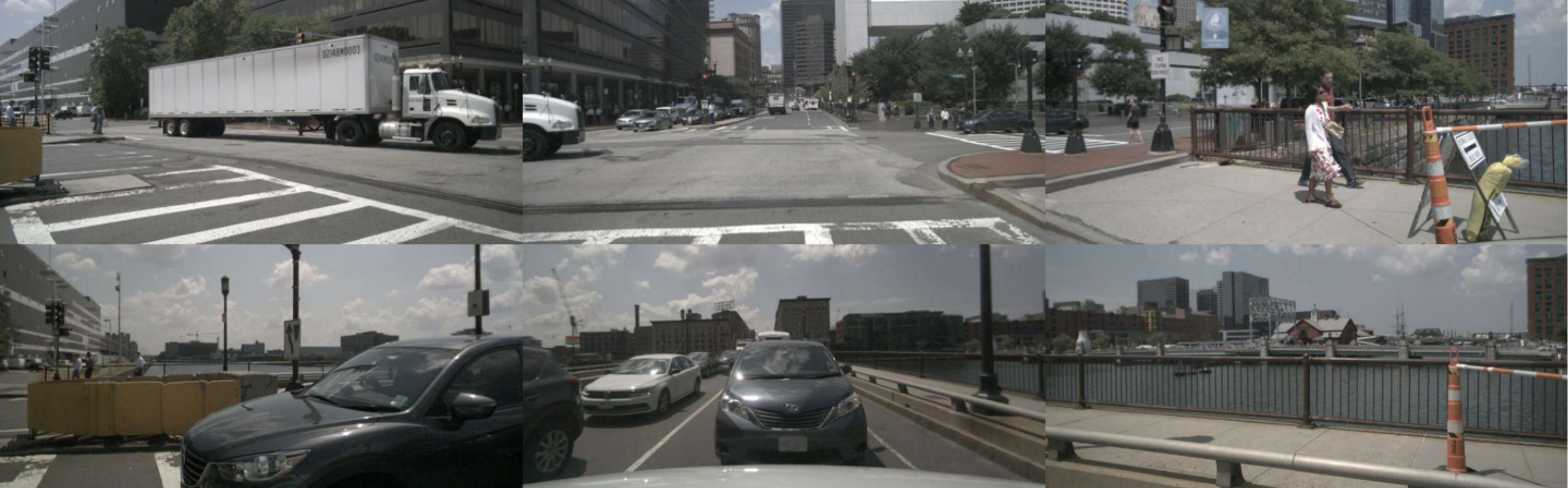}};
    \node[rectangle, camtag, fill=colorFL,  anchor=north west, yshift=-2pt, xshift=50pt] (Cam_FL) at (Cam.north west) {front left};
    \node[rectangle, camtag, fill=colorF, anchor=north, yshift=-2pt] (Cam_F) at (Cam.north) {front};
    \node[rectangle, camtag, fill=colorFR, anchor=north east, yshift=-2pt, xshift=-50pt] (Cam_FR) at (Cam.north east) {front right};
    \node[rectangle, camtag, fill=colorBL, anchor=west, yshift=-5pt, xshift=50pt] (Cam_BL) at (Cam.west) {back left};
    \node[rectangle, camtag, fill=colorB, anchor=center, yshift=-5pt] (Cam_B) at (Cam.center) {back};
    \node[rectangle, camtag, fill=colorBR, anchor=east, yshift=-5pt, xshift=-50pt] (Cam_BR) at (Cam.east) {back right};
\end{tikzpicture}

\caption{Six input camera images coming from the 360-degree camera rig of nuScenes. Note small overlaps between views, e.g., the front of the white truck is both seen in the front-left and front cams.}
\label{fig:attention_study:cam_band}
\end{figure*}